\documentclass[sigconf]{acmart}

\usepackage{xspace}
\usepackage{multirow} 
\usepackage{arydshln}
\usepackage{caption}
\usepackage{subcaption}
\usepackage{makecell}
\usepackage{balance}

\usepackage{tikz}
\newcommand{\myarrow}[1][]{
  \begin{tikzpicture}[#1]
    \draw (0,0.5em) -- (0,0) -- (0.5em,0);
    \draw (0.3em,0.2em) -- (0.5em,0) -- (0.3em,-0.2em);
  \end{tikzpicture}
}

\newtheorem{mydef}{Definition}
\newcommand{\set}[1]{\mathcal{#1}}
\providecommand{\sE}{\ensuremath{\set{E}}}
\providecommand{\sL}{\ensuremath{\set{L}}}

\providecommand{\sP}{\ensuremath{\set{P}}}
\providecommand{\sR}{\ensuremath{\set{R}}}
\providecommand{\sS}{\ensuremath{\set{S}}}
\providecommand{\sT}{\ensuremath{\set{T}}}
\providecommand{\sV}{\ensuremath{\set{V}}}

\renewcommand{\vec}[1]{{\bf{#1}}}
\providecommand{\vb}{\ensuremath{\vec{b}}}

\providecommand{\vh}{\ensuremath{\vec{h}}}
\providecommand{\vm}{\ensuremath{\vec{m}}}
\providecommand{\vp}{\ensuremath{\vec{p}}}
\providecommand{\vq}{\ensuremath{\vec{q}}}
\providecommand{\vr}{\ensuremath{\vec{r}}}

\providecommand{\vt}{\ensuremath{\vec{t}}}
\providecommand{\vu}{\ensuremath{\vec{u}}}
\providecommand{\vecv}{\ensuremath{\vec{v}}}
\providecommand{\vw}{\ensuremath{\vec{w}}}
\providecommand{\vx}{\ensuremath{\vec{x}}}
\providecommand{\vy}{\ensuremath{\vec{y}}}
\providecommand{\vz}{\ensuremath{\vec{z}}}

\newcommand{\mat}[1]{\boldsymbol{#1}}

\providecommand{\mK}{\ensuremath{\mat{K}}}
\providecommand{\mQ}{\ensuremath{\mat{Q}}}
\providecommand{\mV}{\ensuremath{\mat{V}}}
\providecommand{\mW}{\ensuremath{\mat{W}}}
\providecommand{\mX}{\ensuremath{\mat{X}}}
\providecommand{\mZ}{\ensuremath{\mat{Z}}}

\providecommand{\dhat}{\widehat{d}}
\providecommand{\vbhat}{\widehat{\vb}}
\providecommand{\vphat}{\widehat{\vp}}
\providecommand{\mKhat}{\widehat{\mK}}
\providecommand{\mQhat}{\widehat{\mQ}}
\providecommand{\mVhat}{\widehat{\mV}}
\providecommand{\mWhat}{\widehat{\mW}}

\providecommand{\vbbar}{\overline{\vb}}
\providecommand{\vwbar}{\overline{\vw}}
\providecommand{\mWbar}{\overline{\mW}}

\providecommand{\mtX}{\ensuremath{\mat{\widetilde{X}}}}
\providecommand{\mtZ}{\ensuremath{\mat{\widetilde{Z}}}}

\newcommand{\fb}[0]{\textsl{Freebase}\xspace}
\newcommand{\wiki}[0]{\textsl{Wikidata}\xspace}
\newcommand{\yago}[0]{\textsl{YAGO}\xspace}
\newcommand{\yagot}[0]{\textsl{YAGO3}\xspace}
\newcommand{\dbp}[0]{\textsl{DBpedia}\xspace}
\newcommand{\jf}[0]{\textsl{JF17K}\xspace}
\newcommand{\wikip}[0]{\textsl{WikiPeople}\xspace}
\newcommand{\wikipm}[0]{\textsl{WikiPeople$^-$}\xspace}
\newcommand{\wdk}[0]{\textsl{WD50K}\xspace}
\newcommand{\fbk}[0]{\textsl{FB15K237}\xspace}

\newcommand{\hnfb}[0]{\textsl{HN-FB}\xspace}
\newcommand{\hnfbs}[0]{\textsl{HN-FB-S}\xspace}
\newcommand{\hnwk}[0]{\textsl{HN-WK}\xspace}
\newcommand{\hnyg}[0]{\textsl{HN-YG}\xspace}
\newcommand{\hnkg}[0]{\text{HN-KG}\xspace}

\newcommand{\ours}[0]{\textup{HyNT}\xspace}
\newcommand{\mtkgnn}[0]{\textup{MT-KGNN}\xspace}
\newcommand{\transea}[0]{\textup{TransEA}\xspace}

\newcommand{\literale}[0]{\textup{LiteralE}\xspace}
\newcommand{\kblrn}[0]{\textup{KBLRN}\xspace}
\newcommand{\kbln}[0]{\textup{KBLN}\xspace}

\newcommand{\nalp}[0]{\textup{NaLP}\xspace}
\newcommand{\tnalp}[0]{\textup{tNaLP}\xspace}
\newcommand{\neuinfer}[0]{\textup{NeuInfer}\xspace}
\newcommand{\hinge}[0]{\textup{HINGE}\xspace}
\newcommand{\stare}[0]{\textup{StarE}\xspace}
\newcommand{\ram}[0]{\textup{RAM}\xspace}
\newcommand{\hytrans}[0]{\textup{Hy-Transformer}\xspace}
\newcommand{\gran}[0]{\textup{GRAN}\xspace}

\newcommand{\mrr}[0]{MRR\xspace}
\newcommand{\hten}[0]{Hit@10\xspace}
\newcommand{\hthree}[0]{Hit@3\xspace}
\newcommand{\hone}[0]{Hit@1\xspace}
\newcommand{\rmse}[0]{RMSE\xspace}

\AtBeginDocument{%
  \providecommand\BibTeX{{%
    \normalfont B\kern-0.5em{\scshape i\kern-0.25em b}\kern-0.8em\TeX}}}

\copyrightyear{2023}
\acmYear{2023}
\setcopyright{acmlicensed}\acmConference[KDD '23]{Proceedings of the 29th ACM SIGKDD Conference on Knowledge Discovery and Data Mining}{August 6--10, 2023}{Long Beach, CA, USA}
\acmBooktitle{Proceedings of the 29th ACM SIGKDD Conference on Knowledge Discovery and Data Mining (KDD '23), August 6--10, 2023, Long Beach, CA, USA}
\acmPrice{15.00}
\acmDOI{10.1145/3580305.3599490}
\acmISBN{979-8-4007-0103-0/23/08}

\begin{document}

\title{Representation Learning on Hyper-Relational and Numeric Knowledge Graphs with Transformers}

\author{Chanyoung Chung}
\authornote{Authors in alphabetical order with equal contribution.}
\affiliation{%
  \institution{School of Computing, KAIST}
  \city{Daejeon}
  \country{Republic of Korea}
}
\email{chanyoung.chung@kaist.ac.kr}

\author{Jaejun Lee}
\authornotemark[1]
\affiliation{%
  \institution{School of Computing, KAIST}
  \city{Daejeon}
  \country{Republic of Korea}
}
\email{jjlee98@kaist.ac.kr}

\author{Joyce Jiyoung Whang}
\authornote{Corresponding author.}
\affiliation{%
  \institution{School of Computing, KAIST}
  \city{Daejeon}
  \country{Republic of Korea}
}
\email{jjwhang@kaist.ac.kr}


\begin{abstract}
In a hyper-relational knowledge graph, a triplet can be associated with a set of qualifiers, where a qualifier is composed of a relation and an entity, providing auxiliary information for the triplet. While existing hyper-relational knowledge graph embedding methods assume that the entities are discrete objects, some information should be represented using numeric values, e.g., (J.R.R., was born in, 1892). Also, a triplet (J.R.R., educated at, Oxford Univ.) can be associated with a qualifier such as (start time, 1911). In this paper, we propose a unified framework named HyNT that learns representations of a hyper-relational knowledge graph containing numeric literals in either triplets or qualifiers. We define a context transformer and a prediction transformer to learn the representations based not only on the correlations between a triplet and its qualifiers but also on the numeric information. By learning compact representations of triplets and qualifiers and feeding them into the transformers, we reduce the computation cost of using transformers. Using HyNT, we can predict missing numeric values in addition to missing entities or relations in a hyper-relational knowledge graph. Experimental results show that HyNT significantly outperforms state-of-the-art methods on real-world datasets.
\end{abstract}




\maketitle

\section{Introduction}
\label{sec:intro}
\begin{figure}[t]
\centering
\includegraphics[width=\linewidth]{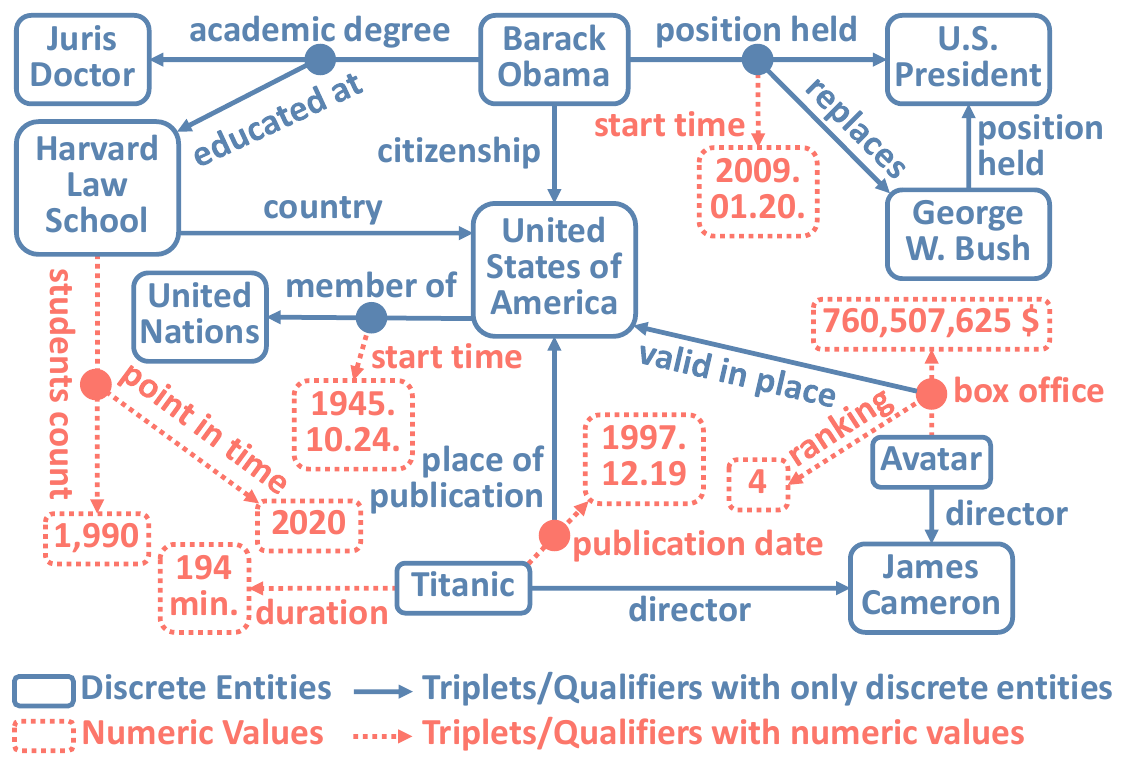}
\caption{A real-world hyper-relational knowledge graph containing numeric literals. This is a subgraph of \hnwk which is created based on \wiki. Details are in Section~\ref{sec:rkg}.}
\label{fig:kg}
\end{figure}

While most research on knowledge graphs assumes a classical form of a knowledge graph composed only of triplets~\cite{survey,surv,smore,meta}, it has been recognized that the triplet format oversimplifies information that can be represented~\cite{hinge}. To enrich information in a knowledge graph, a hyper-relational knowledge graph has been recently studied~\cite{stare,gran}, where a triplet is extended to a hyper-relational fact which is defined by a triplet and its qualifiers. Each qualifier is represented by a relation-entity pair and adds auxiliary information to a triplet. For example, a triplet \textsf{(Barack Obama, academic degree, Juris Doctor)} can be associated with a qualifier \textsf{(educated at, Harvard Law School)} representing that Barack Obama got his JD at Harvard Law School, as shown in Figure~\ref{fig:kg}.


Different hyper-relational knowledge graph embedding methods have been recently proposed, e.g., \hinge~\cite{hinge}, \neuinfer~\cite{neuinfer}, \stare~\cite{stare}, \hytrans~\cite{hytrans}, and \gran~\cite{gran}. However, existing methods treat all entities as discrete objects~\cite{nalp,tnalp,ram,neuinfer,gran} or remove numeric literals~\cite{hinge,stare,hytrans}, even though some information should be represented using numeric values. For example, Figure~\ref{fig:kg} shows a subgraph of a real-world hyper-relational knowledge graph \hnwk (\textsl{H}yper-relational, \textsl{N}umeric \textsl{W}i\textsl{K}idata) which will be described in Section~\ref{sec:rkg}. To represent that the number of students at Harvard Law School was 1,990 in 2020, we have \textsf{(Harvard Law School, students count, 1,990)} and its qualifier \textsf{(point in time, 2020)}.

We propose a representation learning method that learns representations of a hyper-relational knowledge graph containing diverse numeric literals. We name our method \ours which stands for \textbf{Hy}per-relational knowledge graph embedding with \textbf{N}umeric literals using \textbf{T}ransformers. Using well-known knowledge bases, \wiki~\cite{wk}, \yago~\cite{yago}, and \fb~\cite{fb}, we create three real-world datasets that include various numeric literals in triplets and qualifiers. \ours properly encodes and utilizes the numeric literals. By considering a loss that makes the numeric values be recovered using the latent representations, \ours can predict not only missing entities or relations but also missing numeric values. For example, given \textsf{(Titanic, duration, ?)}, \ours can predict ? in Figure~\ref{fig:kg}. Our datasets and implementations are available at \url{https://github.com/bdi-lab/HyNT}. 


\begin{figure}[t]
\centering
\includegraphics[width=0.9\linewidth]{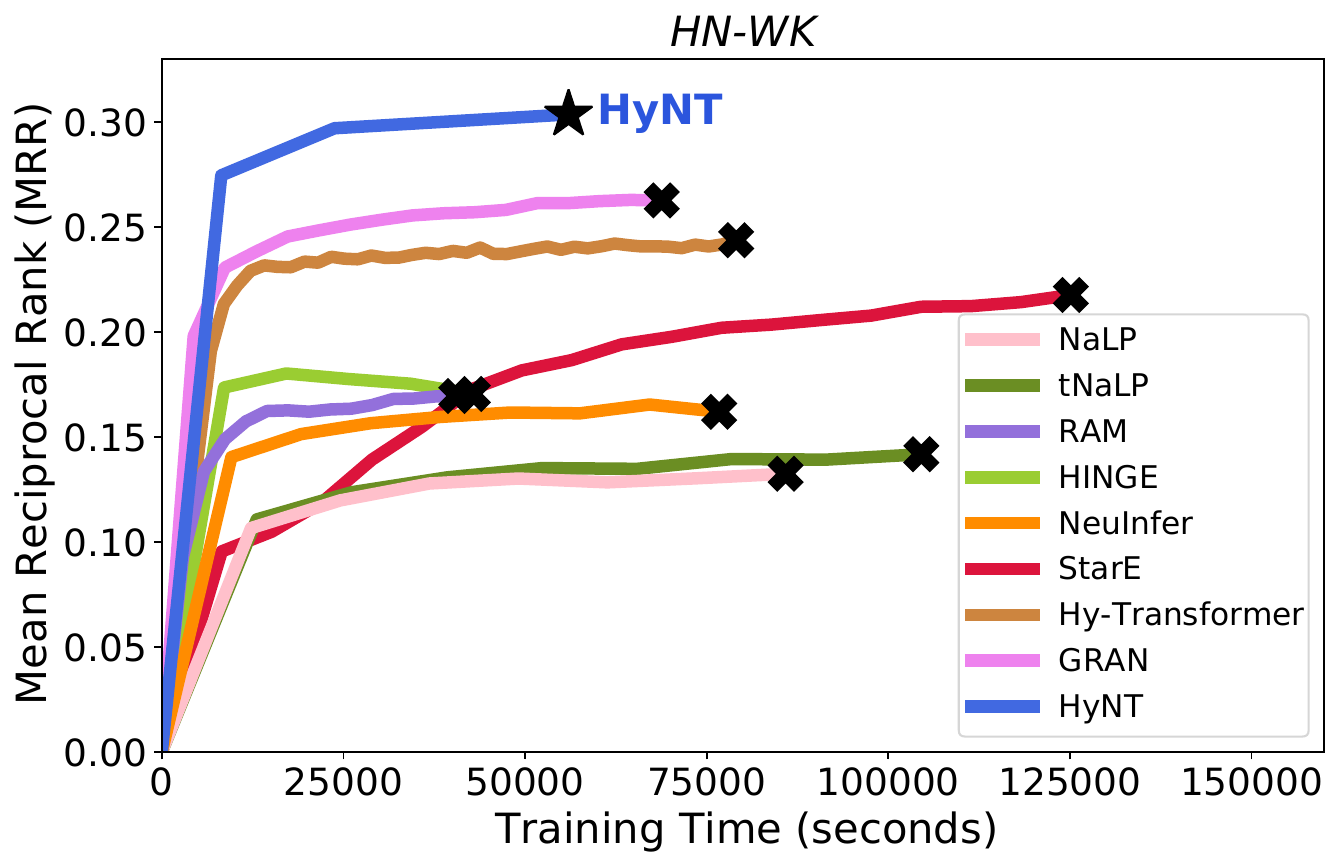}
\caption{Link prediction performance according to the training time on \hnwk. A higher MRR score indicates better performance. \color{black}{\ours performs better than all the other methods while requiring less training time.}}
\label{fig:intro}
\end{figure}

We define a context transformer and a prediction transformer to learn embedding vectors by incorporating not only the structure of a hyper-relational fact but also the information provided by the numeric literals. In particular, \ours learns the representations of triplets and qualifiers by aggregating the embeddings of entities and relations that constitute the corresponding triplets or qualifiers. By feeding the representations of triplets and qualifiers into transformers, \ours effectively captures the correlations between a triplet and its qualifiers while reducing computation cost. Experimental results show that \ours significantly outperforms 12 different baseline methods in link prediction, numeric value prediction, and relation prediction tasks on real-world datasets.

Figure~\ref{fig:intro} shows the Mean Reciprocal Rank (MRR) according to the training time of different methods on \hnwk. MRR is a standard metric for link prediction; the higher, the better. Among the eight baseline methods, the first three methods, \nalp~\cite{nalp}, \tnalp~\cite{tnalp}, and \ram~\cite{ram}, utilize n-ary representations where a hyper-relational fact is represented as role-entity pairs. The rest five methods are hyper-relational knowledge graph embedding methods. We show the MRR score of each method until the method finishes\footnote{We determine each method's best epoch on a validation set and run methods until their best epochs on a test set. In Figure~\ref{fig:intro}, the MRRs are measured on the test set.} or it runs 36 hours. For a fair comparison, we run all methods on GeForce RTX 2080 Ti except \ram because \ram requires more memory. We run \ram using RTX A6000. In Figure~\ref{fig:intro}, we see that \ours {\color{black}{ substantially outperforms all other baseline methods in performance while requiring less training time}}.

\section{Related Work}
\label{sec:related}
\paragraph{{\bf Knowledge Graph Embedding with Numeric Literals}} For normal knowledge graphs but not for hyper-relational knowledge graphs, several methods incorporate numeric literals into knowledge graph embedding~\cite{transea, mtkgnn, kblrn, literale, mkbe, nap, mrap, qleap}. For example, \mtkgnn~\cite{mtkgnn} and \transea~\cite{transea} introduce an additional loss that regresses entity embeddings to their corresponding numeric attributes. \kblrn~\cite{kblrn} collects patterns between numeric attributes, which is used to measure the plausibility of triplets. \literale~\cite{literale} directly utilizes numeric literals to compute entity embeddings using gated recurrent units. However, these models can only handle triplets and cannot appropriately handle hyper-relational facts.

\paragraph{{\bf N-ary Relations \& Hyper-Relational Facts}} Some recent works~\cite{mtransh, rae, nalp, getd, sts, tnalp, ram, hype} extend triplets using n-ary representations where each fact is represented by multiple role-entity pairs. Among them, \nalp~\cite{nalp}, \tnalp~\cite{tnalp}, and \ram~\cite{ram} measure the plausibility of each n-ary fact by modeling the relatedness between role-entity pairs. On the other hand, recent studies use hyper-relational knowledge graphs~\cite{hinge, neuinfer, gran, stare, hytrans, starqe}, pointing out that n-ary representations cannot fully express the diverse information in knowledge bases~\cite{hinge}. \hinge~\cite{hinge} and \neuinfer~\cite{neuinfer} consider the validity of a triplet as well as its qualifiers to measure the plausibility of each hyper-relational fact. Unlike \ours, the aforementioned methods assume that all entities in the n-ary relational or hyper-relational facts are discrete.

\paragraph{{\bf Transformer-based Knowledge Graph Embedding}} Inspired by the great success of transformer-based frameworks~\cite{att, bert}, various methods utilize transformers to solve tasks on knowledge graphs~\cite{hitter, star} and hyper-relational knowledge graphs~\cite{stare, hytrans, gran}. \stare~\cite{stare} uses a GNN-based encoder, which reflects the structure of hyper-relational facts, along with a transformer-based decoder. \hytrans~\cite{hytrans} replaces the GNN-based encoder in \stare with a layer normalization. \gran~\cite{gran} treats each hyper-relational fact as a heterogeneous graph and uses it as an input of a transformer. Different from \ours, these methods do not consider the case of having numeric literals in hyper-relational facts.

\paragraph{{\bf Existing Benchmark Datasets}} Most well-known benchmark datasets for hyper-relational knowledge graphs are \jf~\cite{mtransh,jf}, \wdk~\cite{stare}, \wikipm~\cite{hinge}, and \wikip~\cite{nalp}. When creating \jf, \wdk, and \wikipm, numeric literals were removed because the previous embedding methods could not adequately handle them. On the other hand, the original version of \wikip includes some limited numeric literals, where \wikip was created by extracting facts involving entities of type human from \wiki dump~\cite{nalp}. Thus, the information in \wikip is restricted to humans, and all numeric literals are `year'. To create datasets containing more diverse numeric literals, we make \hnwk using the original \wiki dump without the constraint of considering humans. We also create two more datasets using \yago and \fb. While we focus on evaluating methods on datasets containing various numeric literals, we also provide experimental results on \wikipm and \wdk in Appendix~\ref{app:bench}.

\section{Hyper-Relational Knowledge Graphs Containing Numeric Literals}
We introduce a formal definition of a hyper-relational and numeric knowledge graph and describe real-world datasets and the tasks. 

\subsection{Definition of a Hyper-Relational and Numeric Knowledge Graph}
We assume that an entity in a hyper-relational knowledge graph can be either a discrete object or a numeric literal. If an entity is a discrete object, we call it a discrete entity; if an entity is a numeric literal, we call it a numeric entity. When a hyper-relational knowledge graph includes numeric entities, we call it a Hyper-relational and Numeric Knowledge Graph (\hnkg). When an entity is a numeric entity in \hnkg, the numeric value is accompanied by its unit, e.g., \textsf{80 kg} or \textsf{80 years}. Let $\sV_\mathrm{N}\subset\mathbb{R}\times\sT$ denote a set of numeric entities where $\sT$ denotes a set of units. We formally define an \hnkg as follows:

\begin{mydef}[Hyper-Relational and Numeric Knowledge Graph]
A hyper-relational and numeric knowledge graph is defined by $G = (\sV, \sR, \sE)$ where $\sV$ is a set of entities represented by $\sV\coloneqq\sV_\mathrm{D}\cup\sV_\mathrm{N}$, $\sV_\mathrm{D}$ is a set of discrete entities, $\sV_\mathrm{N}$ is a set of numeric entities, $\sR$ is a set of relations, and $\sE$ is a set of hyper-relational facts defined by $\sE \subset \sE_\mathrm{tri} \times \sP(\sE_\mathrm{qual})$ where $\sE_\mathrm{tri}\subset \sV \times \sR \times \sV$ is a set of primary triplets, $\sE_\mathrm{qual}\subset \sR \times \sV$ is a set of qualifiers, and $\sP(\sS)$ denotes the power set of $\sS$.
\end{mydef}

\subsection{Creating Real-World Datasets}
\label{sec:rkg}
We create three \hnkg datasets: \hnwk, \hnyg, and \hnfb which are created based on \wiki~\cite{wk}, \yago~\cite{yago}, and \fb~\cite{fb}, respectively. Details about how to make these datasets are described in Appendix~\ref{app:data}. According to the standard Resource Description Framework (RDF), the numeric entities should only appear as a tail entity of a primary triplet or a qualifier's entity~\cite{stand}. We classify the numeric entities based on their relations in the primary triplet or qualifier they belong to and make the numeric entities with the same relation have the same unit, e.g., 176.4 lb is converted into 80 kg. We use the International System of Units (SI) in this process, e.g., the mass should be represented in kilograms. By using the same unit per relation, the prediction task can become simpler. Even though a numeric entity consists of its value and unit in general, we can interpret the numeric entities without their units because the semantics of the numeric entities are determined by their relations. For example, let us consider two triplets \textsf{(Robbie Keane, weight, 80 kg)} and \textsf{(Canada, life expectancy, 80 years)}. Even though the numeric values of the tail entities of these two triplets are both \textsf{80}, we can distinguish them by looking at their relations, \textsf{weight} and \textsf{life expectancy}. Therefore, from this point, we drop the units from $\sV_\mathrm{N}$ for brevity, i.e., $\sV_\mathrm{N}\subset\mathbb{R}$. 


\subsection{Link Prediction, Numeric Value Prediction and Relation Prediction Tasks}
\label{sec:task}
On an \hnkg, a link prediction task predicts a missing discrete entity in each hyper-relational fact, whereas a numeric value prediction task predicts a missing numeric value. Also, a relation prediction task predicts a missing relation. 

Consider $G = (\sV, \sR, \sE)$ where $\sE = \{ ((h, r, t), \{ (q_i, v_i) \}_{i=1}^k) : h \in \sV, r \in \sR, t \in \sV, q_i \in \sR, v_i \in \sV \}$ and $k$ is the number of qualifiers in a hyper-relational fact. We define a general entity prediction task on $G$ as predicting a missing component $?$ in one of the following forms: $((?, r, t), \{ (q_i, v_i) \}_{i=1}^k)$, $((h, r, ?), \{ (q_i, v_i) \}_{i=1}^k)$, and $((h, r, t), \{ (q_i, v_i) \}_{i=1, i \ne j}^k \cup \{(q_j, ?)\})$. If the missing component $?$ is a discrete entity, we call it a link prediction task. If the missing component $?$ is a numeric entity, we call it a numeric value prediction task. On the other hand, we define a relation prediction task to be the task of predicting a missing relation $?$ in $((h, ?, t), \{ (q_i, v_i) \}_{i=1}^k)$ or $((h, r, t), \{ (q_i, v_i) \}_{i=1, i \ne j}^k \cup \{(?, v_j)\})$.

\section{Learning on Hyper-Relational and Numeric Knowledge Graphs}
We propose \ours which learns representations of discrete entities, numeric entities, and relations in an \hnkg. 

\subsection{Representations of Triplets and Qualifiers}
\label{sec:aggr}
Given an \hnkg, $G = (\sV, \sR, \sE)$, let us consider a hyper-relational fact represented by $((h, r, t), \{ (q_i, v_i) \}_{i=1}^k)$ where $(h, r, t)$ is a primary triplet and $\{ (q_i, v_i) \}_{i=1}^k$ is a set of qualifiers. Let $d$ denote the dimension of an embedding vector. Each component of the hyper-relational fact is represented as an embedding vector: let $\vh \in \mathbb{R}^d$ denote an embedding vector of $h$, $\vr \in \mathbb{R}^d$ denote an embedding vector of $r$, $\vt \in \mathbb{R}^d$ denote an embedding vector of $t$, $\vq_i \in \mathbb{R}^d$ denote an embedding vector of $q_i$, and $\vecv_i \in \mathbb{R}^d$ denote an embedding vector of $v_i$. We assume all vectors are column vectors.

For the entities, $h$, $t$, and $v_i$, if the entities are discrete entities, we directly learn the corresponding embedding vectors, $\vh$, $\vt$, and $\vecv_i$. On the other hand, if the entities are numeric entities, we utilize their numeric values as well as relation-specific weights and bias vectors to represent the embedding vectors of those numeric entities.\footnote{As discussed in Section~\ref{sec:rkg}, the numeric entities can only be positioned at $t$ or $v_i$.} Specifically, given $(h,r,t)$ with a numeric entity $t\in\sV_\mathrm{N}$, we compute its embedding vector $\vt=t \vw_r + \vb_r$ where $\vw_r \in \mathbb{R}^d$ is a weight vector of a relation $r$ and $\vb_r \in \mathbb{R}^d$ is a bias vector of $r$. Similarly, for a qualifier $(q_i, v_i)$ with $v_i\in\sV_\mathrm{N}$, we compute $\vecv_i=v_i \vw_{q_i} + \vb_{q_i}$ where $\vw_{q_i} \in \mathbb{R}^d$ is a weight vector of a qualifier's relation $q_i$ and $\vb_{q_i} \in \mathbb{R}^d$ is a bias vector of $q_i$.

We define an embedding vector of a primary triplet $(h,r,t)$ as $\vx_\mathrm{tri}$. Using a projection matrix $\mW_\mathrm{tri} \in \mathbb{R}^{d \times 3d}$, we compute $\vx_\mathrm{tri} = \mW_\mathrm{tri} [\vh;\vr;\vt]$ where $[\vu_1;\cdots;\vu_n]$ is a vertical concatenation of vectors $\vu_1, \ldots, \vu_n$. We call this process a triplet encoding. Also, to compute an embedding vector of a qualifier $(q_i, v_i)$, we compute $\vx_{\mathrm{qual}_i} = \mW_\mathrm{qual} [\vq_i;\vecv_i]$ where $\mW_\mathrm{qual} \in \mathbb{R}^{d \times 2d}$ is a projection matrix. We call this process a qualifier encoding.

\subsection{Context Transformer}
\label{sec:ctx}
We define a context transformer to learn representations of the primary triplets and the qualifiers by exchanging information among them. Given a primary triplet $(h,r,t)$ and its qualifiers $\{ (q_i, v_i) \}_{i=1}^k$, the context transformer learns the relative importance of each of the qualifiers to the primary triplet and learns the representation of the primary triplet by aggregating the qualifiers' representations, weighing each of the aggregations using the relative importance. Similarly, the representation of each qualifier is also learned by considering the relative importance of the primary triplet and the other qualifiers to the target qualifier. This process allows the representations of the primary triplet and their qualifiers to reflect the context by considering the correlations between them. 

The input of the context transformer is:
\begin{equation}
\begin{split}
\mX^{(0)}&=[(\vx_\mathrm{tri} + \vp_{\mathrm{tri}})\Vert(\vx_{\mathrm{qual}_1} + \vp_{\mathrm{qual}})\Vert\cdots\Vert(\vx_{\mathrm{qual}_k} + \vp_{\mathrm{qual}})]\\
&=[\vx_\mathrm{tri}^{(0)}\Vert\vx_{\mathrm{qual}_1}^{(0)}\Vert\cdots\Vert\vx_{\mathrm{qual}_k}^{(0)}]
\end{split}
\end{equation}
where $\vx_\mathrm{tri}$ is the embedding of a primary triplet and $\vx_{\mathrm{qual}_i}$ is the embedding of the $i$-th qualifier, respectively, discussed in Section~\ref{sec:aggr}, $\vp_{\mathrm{tri}}\in \mathbb{R}^d$ is a learnable vector for positional encoding of a primary triplet for the context transformer, $\vp_{\mathrm{qual}} \in \mathbb{R}^d$ is another learnable vector for positional encoding of a qualifier for the context transformer, and $[\vu_1\Vert\cdots\Vert\vu_n]$ is a horizontal concatenation of vectors $\vu_1, \ldots, \vu_n$. By introducing $\vp_{\mathrm{tri}}$ and $\vp_{\mathrm{qual}}$, we encode the information about whether the corresponding embedding vector indicates a primary triplet or one of its qualifiers.

In the attention layer, we compute
\begin{displaymath}
\mtX^{(l)} = \mV^{(l)} \mX^{(l)} \mathrm{softmax}\left(\frac{(\mQ^{(l)} \mX^{(l)})^T(\mK^{(l)} \mX^{(l)})}{\sqrt{d}}\right)
\end{displaymath}
where $\mQ^{(l)}, \mK^{(l)}, \mV^{(l)} \in \mathbb{R}^{d \times d}$ are projection matrices for query, key, and value, respectively~\cite{att}, in the context transformer with $l=0,\cdots,L_\mathrm{C}-1$ and $L_\mathrm{C}$ is the number of layers in the context transformer. We use the multi-head attention mechanism with $n_\mathrm{C}$ heads where $n_\mathrm{C}$ is a hyperparameter~\cite{att,gat}. Then, we employ a residual connection~\cite{add} and apply layer normalization~\cite{norm}. The following feedforward layer is computed by 
\begin{displaymath}
\mX^{(l+1)} = \mW_2^{(l)}\sigma\left(\mW_1^{(l)} \mtX^{(l)} + \vb_1^{(l)} \right) + \vb_2^{(l)}
\end{displaymath}
where $\sigma(x)=\max(0, x)$ is the ReLU function, $\mW_1^{(l)} \in \mathbb{R}^{d_\mathrm{F} \times d}$, $\mW_2^{(l)} \in \mathbb{R}^{d \times d_\mathrm{F}}$, $\vb_1^{(l)} \in \mathbb{R}^{d_\mathrm{F}}$,  $\vb_2^{(l)} \in \mathbb{R}^d$, and $d_\mathrm{F}$ is the hidden dimension of the feedforward layer in the context transformer. Then, we again employ a residual connection, followed by layer normalization. By repeating the above process for $l=0,\cdots,L_\mathrm{C}-1$, we get the final representations of a primary triplet and its qualifiers: $\mX^{(L_\mathrm{C})}=[\vx_\mathrm{tri}^{(L_\mathrm{C})}\Vert \vx_{\mathrm{qual}_1}^{(L_\mathrm{C})}\Vert \cdots\Vert \vx_{\mathrm{qual}_k}^{(L_\mathrm{C})}]$ for each hyper-relational fact.

\subsection{Prediction Transformer}
We define a prediction transformer that learns representations used to predict a missing component in a primary triplet or predict a missing component in a qualifier. To make a prediction on a primary triplet $(h,r,t)$, the input of the prediction transformer is defined by
\begin{equation}
\begin{split}
\mZ^{(0)} &= [(\vx_\mathrm{tri}^{(L_\mathrm{C})} + \vphat_\mathrm{tri})\Vert(\vh + \vphat_\mathrm{h})\Vert(\vr + \vphat_\mathrm{r})\Vert(\vt + \vphat_\mathrm{t})]\\
&=[\vz_{\mathrm{tri}}^{(0)}\Vert \vh^{(0)}\Vert \vr^{(0)}\Vert \vt^{(0)}]
\end{split}
\end{equation}where $\vx_\mathrm{tri}^{(L_\mathrm{C})}$ is the representation of the primary triplet returned by the context transformer, $\vh$, $\vr$, and $\vt$ are the embedding vectors of $h$, $r$, and $t$, respectively, and $\vphat_\mathrm{tri}$, $\vphat_\mathrm{h}$, $\vphat_\mathrm{r}$, $\vphat_\mathrm{t}\in\mathbb{R}^d$ are learnable vectors for positional encoding of the primary triplet, a head entity, a relation, and a tail entity, respectively.

To make a prediction on a qualifier $(q_i,v_i)$, the input of the prediction transformer is defined by
\begin{equation}
\begin{split}
\mZ^{(0)} &= [(\vx_{\mathrm{qual}_i}^{(L_\mathrm{C})} + \vphat_\mathrm{qual})\Vert(\vq_i + \vphat_\mathrm{q})\Vert(\vecv_i + \vphat_\mathrm{v})]\\
&=[\vz_{\mathrm{qual}_i}^{(0)}\Vert {\vq_i}^{(0)}\Vert {\vecv_i}^{(0)}]
\end{split}
\end{equation}where $\vx_{\mathrm{qual}_i}^{(L_\mathrm{C})}$ is the representation of the qualifier returned by the context transformer, $\vq_i$ and $\vecv_i$ are the embedding vectors of $q_i$ and $v_i$, respectively, and $\vphat_\mathrm{qual}$, $\vphat_\mathrm{q}$, $\vphat_\mathrm{v}\in\mathbb{R}^d$ are learnable vectors for positional encoding of a qualifier, a relation in a qualifier, and an entity in a qualifier, respectively.

In the attention layer, we compute
\begin{displaymath}
\mtZ^{(l)} = \mVhat^{(l)} \mZ^{(l)} \mathrm{softmax}\left(\frac{(\mQhat^{(l)} \mZ^{(l)})^T(\mKhat^{(l)} \mZ^{(l)})}{\sqrt{d}}\right)
\end{displaymath}
where $\mQhat^{(l)}, \mKhat^{(l)}, \mVhat^{(l)} \in \mathbb{R}^{d \times d}$ are projection matrices for query, key, and value, respectively, in the prediction transformer with $l=0,\cdots,L_\mathrm{P}-1$, and $L_\mathrm{P}$ is the number of layers in the prediction transformer. We use the multi-head attention mechanism with $n_\mathrm{P}$ heads where $n_\mathrm{P}$ is a hyperparameter. After employing a residual connection and layer normalization, the feedforward layer is defined by\
\begin{displaymath}
\mZ^{(l+1)} = \mWhat_2^{(l)}\sigma\left( \mWhat_1^{(l)} \mtZ^{(l)} + \vbhat_1^{(l)} \right) + \vbhat_2^{(l)}
\end{displaymath}
where $\mWhat_1^{(l)} \in \mathbb{R}^{\dhat_\mathrm{F} \times d}$, $\mWhat_2^{(l)} \in \mathbb{R}^{d \times \dhat_\mathrm{F}}$, $\vbhat_1^{(l)} \in \mathbb{R}^{\dhat_\mathrm{F}}$, $\vbhat_2^{(l)} \in \mathbb{R}^d$, and $\dhat_\mathrm{F}$ is the hidden dimension of the feedforward layer in the prediction transformer. Then, we again apply the residual connection and layer normalization.

By repeating the above process for $l=0,\cdots,L_\mathrm{P}-1$, we get the representation:
\begin{displaymath}
\mZ^{(L_\mathrm{P})}=[\vz_{\mathrm{tri}}^{(L_\mathrm{P})}\Vert \vh^{(L_\mathrm{P})}\Vert \vr^{(L_\mathrm{P})}\Vert \vt^{(L_\mathrm{P})}]
\end{displaymath} for the case of prediction on a primary triplet, and
\begin{displaymath}
\mZ^{(L_\mathrm{P})}=[\vz_{\mathrm{qual}_i}^{(L_\mathrm{P})}\Vert {\vq_i}^{(L_\mathrm{P})}\Vert {\vecv_i}^{(L_\mathrm{P})}]
\end{displaymath} for the case of prediction on a qualifier.

\subsection{Training of \ours}
\label{sec:loss}
To train \ours, we adopt a masking strategy~\cite{bert}; we introduce three different types of masks depending on the prediction tasks.

\begin{figure}[t]
\centering
\begin{subfigure}{\linewidth}
\centering
\includegraphics[width = \textwidth]{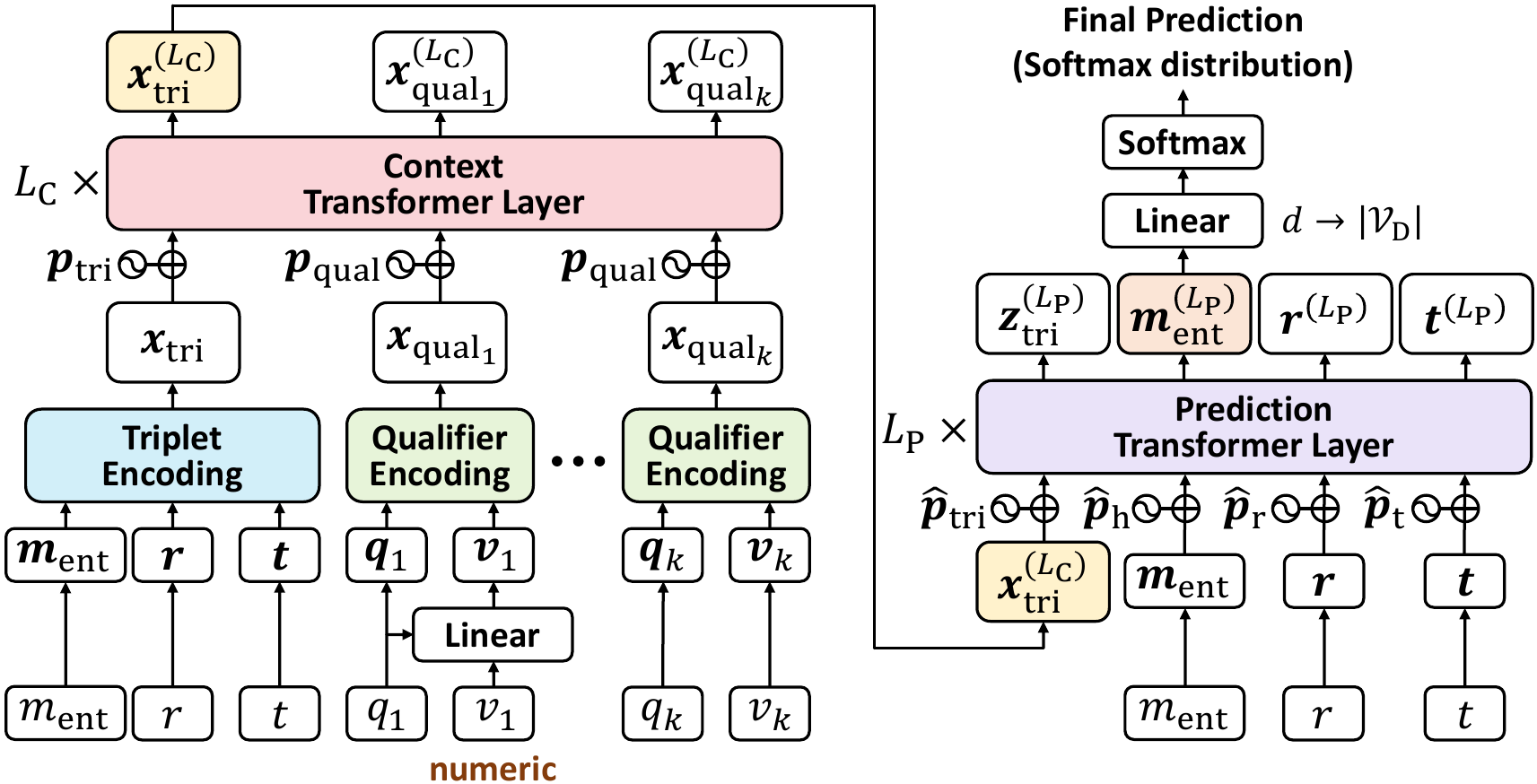}
\caption{Predicting a discrete head entity in a primary triplet.}
\end{subfigure}
\begin{subfigure}{\linewidth}
\centering
\includegraphics[width = \textwidth]{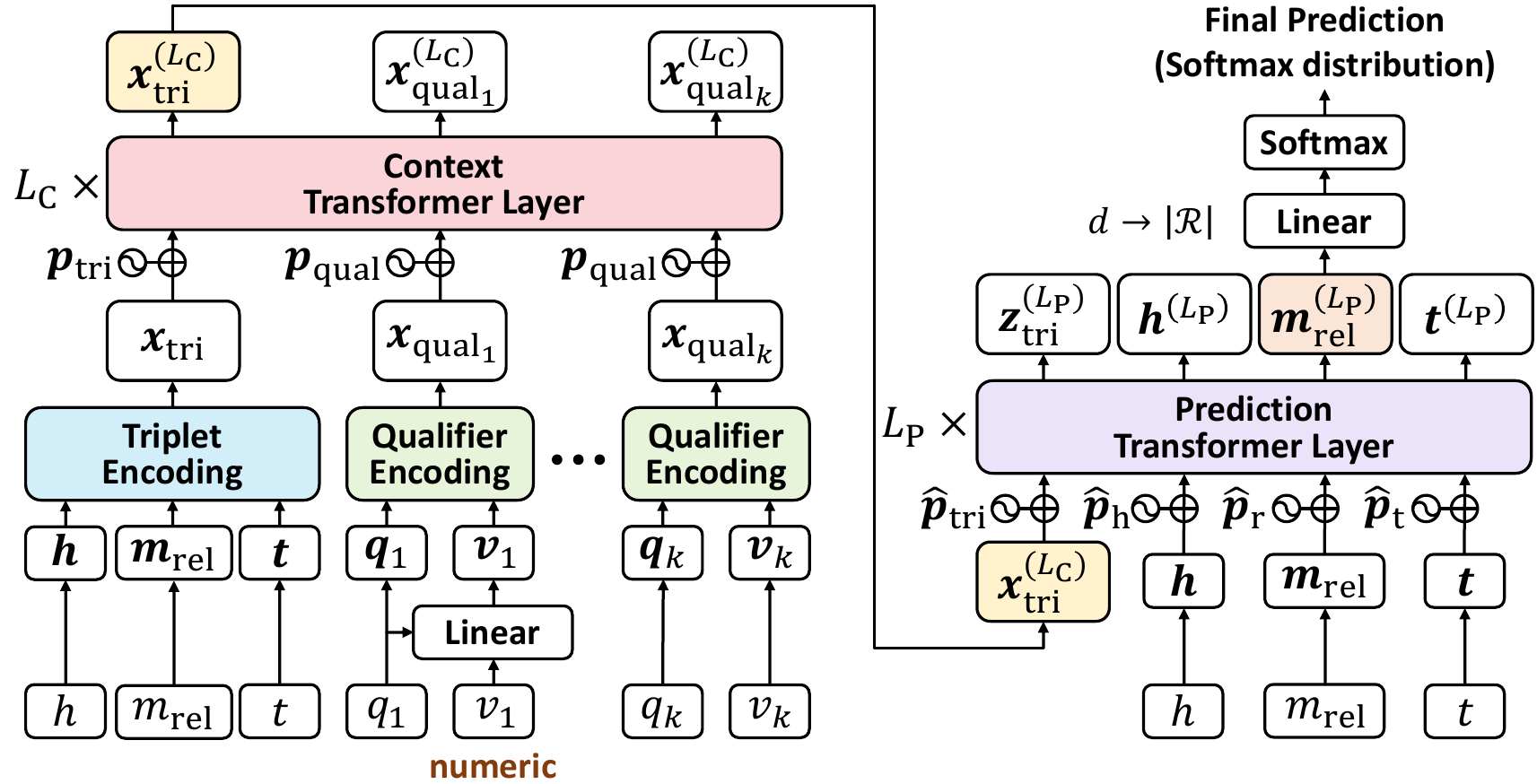}
\caption{Predicting a relation in a primary triplet.}
\end{subfigure}
\begin{subfigure}{\linewidth}
\centering
\includegraphics[width = \textwidth]{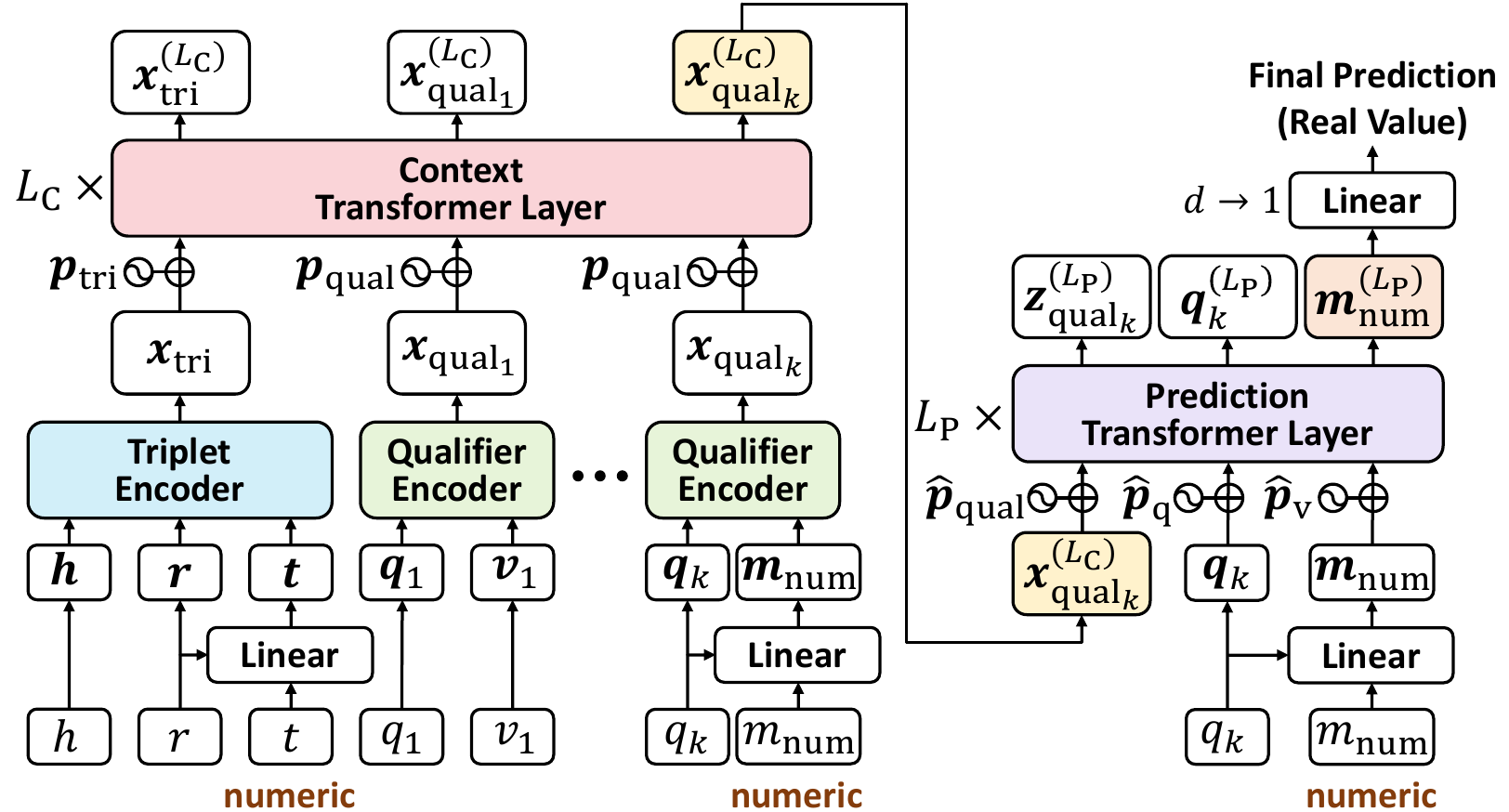}
\caption{Predicting a numeric value in a qualifier.}
\end{subfigure}
\caption{Examples of the training procedure of \ours. We replace a missing component with a mask and train the model to recover the missing component to the ground-truth one.}
\label{fig:train}
\end{figure}

\subsubsection{Discrete Entity Prediction Loss}
We consider the task of predicting a discrete entity in a hyper-relational fact. We introduce a special entity for a mask, denoted by $m_\mathrm{ent}$, and consider it as one of the discrete entities. This special entity $m_\mathrm{ent}$ is associated with a learnable embedding vector denoted by $\vm_\mathrm{ent} \in \mathbb{R}^d$. Given a hyper-relational fact in a training set, we replace one of the discrete entities in the given fact with $m_\mathrm{ent}$ and train the model so that the replaced entity can be recovered. Let $\vm_\mathrm{ent}^{(L_\mathrm{P})} \in \mathbb{R}^d$ be the output representation of the prediction transformer of $m_\mathrm{ent}$. Using a projection matrix $\mWbar_{\mathrm{ent}} \in \mathbb{R}^{|\sV_\mathrm{D}| \times d}$ and a bias vector $\vbbar_{\mathrm{ent}} \in \mathbb{R}^{|\sV_\mathrm{D}|}$, we compute a probability distribution $\vy\in\mathbb{R}^{|\sV_\mathrm{D}|}$ by $\vy = \mathrm{softmax}(\mWbar_{\mathrm{ent}}\vm_\mathrm{ent}^{(L_\mathrm{P})} + \vbbar_{\mathrm{ent}})$ where the $j$-th element of $\vy$ is the probability that the masked entity $m_\mathrm{ent}$ is the $j$-th entity in $\sV_\mathrm{D}$. To compute the loss incurred by the matching between the masked entity and the ground-truth entity, we use the cross entropy loss: $\sL_{\mathrm{ent}} \coloneqq -\log y_j$ where $j$ is the index of the masked entity and $y_j$ is the $j$-th element of $\vy$. In Figure~\ref{fig:train}(a), we show the training procedure of the discrete entity prediction task where we mask the head entity in a primary triplet and train the model so that the masked entity can be recovered.

\subsubsection{Relation Prediction Loss}
We consider the task of predicting a relation. We define a special relation for a mask denoted by $m_\mathrm{rel}$ with its learnable embedding vector $\vm_\mathrm{rel} \in \mathbb{R}^d$. Given a hyper-relational fact in a training set, we replace one of the relations with $m_\mathrm{rel}$. Let $\vm_\mathrm{rel}^{(L_\mathrm{P})} \in \mathbb{R}^d$ be the output representation of the prediction transformer of $m_\mathrm{rel}$. Using $\mWbar_{\mathrm{rel}} \in \mathbb{R}^{|\sR| \times d}$ and $\vbbar_{\mathrm{rel}} \in \mathbb{R}^{|\sR|}$, we compute $\vy = \mathrm{softmax}(\mWbar_{\mathrm{rel}}\vm_\mathrm{rel}^{(L_\mathrm{P})} + \vbbar_{\mathrm{rel}})$ where the $j$-th element indicates the probability that the masked relation is the $j$-th relation in $\sR$. To compute the relation prediction loss, we compute $\sL_{\mathrm{rel}} \coloneqq -\log y_j$ where $j$ is the index of the masked relation and $y_j$ is the $j$-th element of $\vy$. In Figure~\ref{fig:train}(b), we show an example where we mask the relation in a primary triplet and train the model so that the masked relation can be recovered.

\subsubsection{Numeric Value Prediction Loss}
We consider the task of predicting a numeric value. We define a special learnable parameter $m_\mathrm{num} \in \mathbb{R}$ for masking and replace one of the numeric entities with $m_\mathrm{num}$ in a hyper-relational fact. Let $\vm_\mathrm{num}^{(L_\mathrm{P})} \in \mathbb{R}^d$ be the output representation of the prediction transformer of $m_\mathrm{num}$ and $r_m \in \sR$ be the relation associated with the masked numeric value. Using $\vwbar_{r_m} \in \mathbb{R}^d$ and $\overline{b}_{r_m} \in \mathbb{R}$, we predict the masked value by $v_\mathrm{pred} = \vwbar_{r_m} \cdot \vm_\mathrm{num}^{(L_\mathrm{P})} + \overline{b}_{r_m}$. The loss of this prediction is computed by $\sL_{\mathrm{num}} \coloneqq (v_\mathrm{gt} - v_\mathrm{pred})^2$ where $v_\mathrm{gt}$ is the ground-truth masked numeric value. In Figure~\ref{fig:train}(c), we show the numeric value prediction task where we mask a numeric value of a qualifier and train the model so that the ground-truth numeric value can be recovered.

\subsubsection{Joint Loss \& Implementation Details}
We define the final loss function of \ours by adding all three aforementioned losses with appropriate weights: $\sL \coloneqq \sL_\mathrm{ent} + \lambda_1 \cdot \sL_\mathrm{rel} + \lambda_2 \cdot \sL_\mathrm{num}$ where $\lambda_1$ and $\lambda_2$ are hyperparameters governing the relative importance of the relation prediction loss and the numeric value prediction loss, respectively. In our experiments, we set $\lambda_1=\lambda_2=1$.

We implement \ours using PyTorch~\cite{torch}. In our implementation and the experiments, we fix the seeds for random number generators for reproducibility. To prevent overfitting, we apply the label smoothing~\cite{smooth} on $\sL_\mathrm{ent}$ and $\sL_\mathrm{rel}$ with the label smoothing ratio $\epsilon$. We also apply the dropout strategy~\cite{drop} with the dropout rate $\delta$ in both the context transformer and the prediction transformer layers. We utilize the Adam Algorithm~\cite{adam} and the cosine annealing learning rate scheduler with restarts~\cite{cos}.

\subsection{Predictions using \ours}
We describe how \ours performs the link prediction, relation prediction, and numeric value prediction tasks.

\subsubsection{Link Prediction Using \ours}
Consider a link prediction task, $((?, r, t), \{ (q_i, v_i) \}_{i=1}^k)$. To predict the missing discrete entity ?, we replace ? with $m_\mathrm{ent}$ in the given hyper-relational fact and use it as the input of our model. Using the output representation $\vm_\mathrm{ent}^{(L_\mathrm{P})}$ of $m_\mathrm{ent}$ from the prediction transformer, we calculate the probability distribution $\vy = \mathrm{softmax}(\mWbar_{\mathrm{ent}}\vm_\mathrm{ent}^{(L_\mathrm{P})} + \vbbar_{\mathrm{ent}})$. We predict the missing entity ? to be the entity with the highest probability in $\vy$. As discussed in Section~\ref{sec:task}, the position of the missing entity ? can be changed to a tail entity of a primary triplet or a qualifier's entity. 


\subsubsection{Relation Prediction Using \ours}
Given a relation prediction task, $((h, ?, t), \{ (q_i, v_i) \}_{i=1}^k)$, we replace ? with $m_\mathrm{rel}$. Using the output representation $\vm_\mathrm{rel}^{(L_\mathrm{P})}$ of $m_\mathrm{rel}$ from the prediction transformer, we calculate $\vy = \mathrm{softmax}(\mWbar_{\mathrm{rel}}\vm_\mathrm{rel}^{(L_\mathrm{P})} + \vbbar_{\mathrm{rel}})$ and predict ? to be the relation with the highest probability in $\vy$. The position of ? can be changed to a qualifier's relation. 

\subsubsection{Numeric Value Prediction Using \ours}
In a hyper-relational fact $((h, r, t), \{ (q_i, v_i) \}_{i=1, i \ne j}^k \cup \{(q_j, ?)\})$, we replace ? with $m_\mathrm{num}$. Using the output representation $\vm_\mathrm{num}^{(L_\mathrm{P})}$ of $m_\mathrm{num}$ from the prediction transformer, we predict the missing value by $v_\mathrm{pred} = \vwbar_{q_j} \cdot \vm_\mathrm{num}^{(L_\mathrm{P})} + \overline{b}_{q_j}$ where $v_\mathrm{pred}$ is our prediction. Similarly, we can also predict a numeric value in a primary triplet.

\section{Experimental Results}
We compare the performance of \ours with other state-of-the-art methods using real-world \hnkg datasets.

\subsection{Datasets, Baseline Methods, and the Settings}
\label{sec:base}
We use three real-world {\hnkg}s described in Section~\ref{sec:rkg}. Table~\ref{tb:kg} shows the statistic of these datasets where $|\sR_\mathrm{D}|$ is the number of relations only having discrete entities, $|\sR_\mathrm{N}|$ is the number of relations involving numeric entities, $|\sE_\mathrm{tri\_D}|$ is the number of primary triplets that consist of only discrete entities, $|\sE_\mathrm{tri\_N}|$ is the number of primary triplets that include numeric entities, $|\sE_\mathrm{qual\_D}|$ is the number of qualifiers containing discrete entities, and $|\sE_\mathrm{qual\_N}|$ is the number of qualifiers containing numeric entities. We randomly split $\sE$ into training, validation, and test sets with a ratio of 8:1:1. We also show the number of triplets associated with or without qualifiers, denoted by w/ qual. and w/o qual., respectively.

We compare the performance of \ours with 12 baseline methods: \transea~\cite{transea}, \mtkgnn~\cite{mtkgnn}, \kbln~\cite{kblrn}, \literale~\cite{literale}, \nalp~\cite{nalp}, \tnalp~\cite{tnalp}, \ram~\cite{ram}, \hinge~\cite{hinge}, \neuinfer~\cite{neuinfer}, \stare~\cite{stare}, \hytrans~\cite{hytrans}, and \gran~\cite{gran}. The first four methods can handle numeric literals in knowledge graphs and cannot handle hyper-relational facts. On the other hand, the last eight methods handle hyper-relational knowledge graphs and do not consider numeric literals. Note that \stare, \hytrans, and \gran are transformer-based methods. Among the baseline methods, \nalp, \tnalp, \ram, \hinge, \neuinfer, \stare, \hytrans, and \gran failed to process \hnfb due to scalability issues. Since the maximum number of qualifiers attached to a primary triplet is 358 in \hnfb, these methods require much more memory than any machines that we have. Thus, we create a smaller version, \hnfbs, by restricting the maximum number of qualifiers to five.\footnote{We show the results of \ours on \hnfb in Appendix~\ref{app:fb}. Note that \ours is the only method that can handle \hnfb among the methods considering hyper-relational facts.} Also, these eight baseline methods cannot handle numeric literals. To feed our datasets into these methods, we treat the numeric entities as discrete entities by following how the authors of \nalp, \tnalp, \ram, \neuinfer, and \gran run their methods on datasets containing numeric literals. When these methods predict missing numeric entities, e.g., $(h,r,?)$ and ? is a numeric entity, we narrow down the candidates for ? by filtering out the entities that never appear as a tail entity or a qualifier's entity of $r$. For example, to solve \textsf{(Robbie Keane, weight, ?)}, the candidates can be \textsf{80 kg} and \textsf{75 kg} as long as they have appeared as a tail entity or a qualifier's entity of \textsf{weight} in a training set; but \textsf{120 years} is filtered out and cannot be considered as a candidate for~?. We set $d=256$ for all methods and all datasets except for \ram on \hnwk. For \ram, we set $d=64$ on \hnwk due to an out-of-memory issue. Details about how to run the baseline methods are described in Appendix~\ref{app:base}.


\begin{table}[t]
\caption{Real-world \hnkg Datasets.}
\setlength{\tabcolsep}{0.42em}
\label{tb:kg}
\begin{tabular}{clrrrr}
\toprule
 & & \hnwk & \hnyg & \hnfb & \hnfbs \\
\midrule
\multirow{3}{*}{$\sV$} & $|\sV_\mathrm{D}|$ & 13,655 & 12,439 & 14,284 & 5,510\\
 & $|\sV_\mathrm{N}|$ & 79,600 & 24,508 & 62,056 & 21,824 \\
 & $|\sV|$ & 93,255 & 36,947 & 76,340 & 27,334 \\
\cdashline{1-6}
\multirow{3}{*}{$\sR$} & $|\sR_\mathrm{D}|$ & 200 & 31 & 278 & 161 \\
 & $|\sR_\mathrm{N}|$ & 157 & 26 & 83 & 47 \\
 & $|\sR|$ & 357 & 57 & 361 & 208 \\
\cdashline{1-6}
\multirow{3}{*}{$\sE$} & $|\sE|$ & 296,783 & 76,383 & 284,288 & 108,140 \\
 & \myarrow \ w/ qual. & 126,524 & 9,916 & 71,598 & 30,121 \\
 & \myarrow \ w/o qual. & 170,259 & 66,467 & 212,690 & 78,019 \\
\cdashline{1-6}
\multirow{3}{*}{$\sE_\mathrm{tri}$} & $|\sE_\mathrm{tri\_D}|$ & 155,394 & 45,001 & 185,269 & 73,686 \\
 & $|\sE_\mathrm{tri\_N}|$ & 112,787 & 31,382 & 74,017 & 25,701 \\
 & $|\sE_\mathrm{tri}|$ & 268,181 & 76,383 & 259,286 & 99,387 \\
\cdashline{1-6}
\multirow{3}{*}{$\sE_\mathrm{qual}$} & $|\sE_\mathrm{qual\_D}|$ & 3,894 & 4 & 9,416 & 3,775 \\
 & $|\sE_\mathrm{qual\_N}|$ & 9,218 & 8,008 & 2,639 & 500 \\
 & $|\sE_\mathrm{qual}|$ & 13,112 & 8,012 & 12,055 & 4,275 \\
\bottomrule
\end{tabular}
\end{table}

\begin{table*}[t]
\caption{Link Prediction Results on the Primary Triplets in \hnwk, \hnyg, and \hnfbs. The best results are boldfaced and the second-best results are underlined. Our model, \ours, significantly outperforms all baseline methods in terms of all metrics.}
\setlength{\tabcolsep}{0.56em}
\label{tb:lp_tri}
\begin{tabular}{c|cccc|cccc|cccc}
\Xhline{2\arrayrulewidth}
 & \multicolumn{4}{c|}{\hnwk} & \multicolumn{4}{c|}{\hnyg} & \multicolumn{4}{c}{\hnfbs}\\
 & \mrr & \hten & \hthree & \hone & \mrr & \hten & \hthree & \hone & \mrr & \hten & \hthree & \hone \\
\Xhline{\arrayrulewidth}
\transea & 0.1413 & 0.2921 & 0.1728 & 0.0613 & 0.1397 & 0.2707 & 0.1738 & 0.0660 & 0.2565 & 0.4647 & 0.2912 & 0.1554 \\
\mtkgnn & 0.1448 & 0.2244 & 0.1566 & 0.1035 & 0.1377 & 0.2181 & 0.1474 & 0.0998 & 0.2509 & 0.4383 & 0.2758 & 0.1612 \\
\kbln & 0.1500 & 0.2275 & 0.1620 & 0.1112 & 0.1540 & 0.2231 & 0.1690 & 0.1161 & 0.2406 & 0.4303 & 0.2586 & 0.1528 \\
\literale & 0.1635 & 0.2563 & 0.1792 & 0.1182 & 0.1577 & 0.2459 & 0.1733 & 0.1133 & 0.2632 & 0.4612 & 0.2885 & 0.1701 \\
\nalp & 0.1326 & 0.2303 & 0.1359 & 0.0771 & 0.0838 & 0.1529 & 0.0864 & 0.0489 & 0.3721 & 0.5602 & 0.4283 & 0.2726 \\
\tnalp & 0.1419 & 0.2446 & 0.1460 & 0.0859 & 0.1074 & 0.1790 & 0.1163 & 0.0690 & 0.3410 & 0.5165 & 0.3899 & 0.2499 \\
\ram & 0.1696 & 0.2986 & 0.1746 & 0.1032 & 0.1682 & 0.2641 & 0.1862 & 0.1185 & \underline{0.5077} & 0.6269 & 0.5402 & \color{black}{\textbf{0.4432}} \\
\hinge & 0.1706 & 0.2880 & 0.1770 & 0.1051 & 0.1493 & 0.2463 & 0.1640 & 0.1003 & 0.4147 & 0.6262 & 0.4729 & 0.3060 \\
\neuinfer & 0.1621 & 0.2651 & 0.1707 & 0.1052 & 0.1213 & 0.2042 & 0.1327 & 0.0774 & 0.2872 & 0.4725 & 0.3297 & 0.1943 \\
\stare & 0.2177 & 0.3523 & 0.2268 & 0.1438 & 0.1826 & 0.2907 & 0.2042 & 0.1262 & 0.4707 & 0.6521 & 0.5254 & 0.3734 \\
\hytrans & 0.2438 & 0.3463 & 0.2567 & 0.1812 & 0.1884 & 0.2968 & 0.2070 & \underline{0.1335} & 0.4911 & \underline{0.6577} & 0.5410 & 0.4012 \\
\gran & \underline{0.2627} & \underline{0.3761} & \underline{0.2738} & \underline{0.2029} & \underline{0.1951} & \underline{0.3137} & \underline{0.2223} & 0.1319 & 0.5028 & 0.6570 & \underline{0.5488} & \color{black}{\underline{0.4203}} \\
\ours & \color{black}{\textbf{0.3037}} & \color{black}{\textbf{0.5082}} & \color{black}{\textbf{0.3228}} & \color{black}{\textbf{0.2084}} & \color{black}{\textbf{0.2035}} & \color{black}{\textbf{0.3147}} & \color{black}{\textbf{0.2237}} & \color{black}{\textbf{0.1474}} & \color{black}{\textbf{0.5079}} & \color{black}{\textbf{0.7037}} & \color{black}{\textbf{0.5610}} & \color{black}{0.4063} \\
\Xhline{2\arrayrulewidth}
\end{tabular}
\end{table*}

\begin{table*}[t]
\caption{Link Prediction Results of \ours on \hnwk. The predictions made by \ours are changed depending on the qualifiers. \ours successfully predicts the missing entities in the primary triplets by considering their qualifiers.}
\label{tb:lp_qual}
\begin{tabular}{l@{\hskip 7.7em}c}
\toprule
Link Prediction Problem & Prediction \\ 
\midrule
\textsf{((?, nominated for, Best Actor), $\{$(for work, Moneyball), (subject of, 84th Academy Awards)$\}$)} & \textsf{Brad Pitt} \\
\textsf{((?, nominated for, Best Actor), $\{$(for work, Forrest Gump), (subject of, 67th Academy Awards)$\}$)} & \textsf{Tom Hanks} \\
\cdashline{1-2}
\textsf{((?, diplomatic relation, Nicaragua), $\{$(start time, 1979-08-21)$\}$)} & \textsf{North Korea} \\
\textsf{((?, diplomatic relation, Nicaragua), $\{$(start time, 1985-12-07), (end time, 1990-11-09)$\}$)} & \textsf{China} \\
\cdashline{1-2}
\textsf{((London, country, ?), $\{$(start time, 927), (end time, 1707-04-30)$\}$)} & \textsf{Kingdom of England} \\
\textsf{((London, country, ?), $\{$(start time, 1922-12-06)$\}$)} & \textsf{United Kingdom} \\
\cdashline{1-2}
\textsf{((Best Actress, winner, ?), $\{$(for work, The Iron Lady), (point in time, 2011)$\}$)} & \textsf{Meryl Streep}\\
\textsf{((Best Actress, winner, ?), $\{$(for work, The Hours), (point in time, 2002)$\}$)} & \textsf{Nicole Kidman}\\
\bottomrule
\end{tabular}
\end{table*}

\begin{table}[t]
\caption{Link Prediction Results on All Entities of Hyper-Relational Facts in \hnwk and \hnfbs. }
\setlength{\tabcolsep}{0.4em}
\label{tb:lp_all}
\begin{tabular}{cccccc}
\toprule
 & & \mrr & \hten & \hthree & \hone \\
\midrule
\multirow{7}{*}{\hnwk} & \nalp & 0.1541 & 0.2560 & 0.1602 & 0.0965 \\
 & \tnalp & 0.1616 & 0.2693 & 0.1685 & 0.1030 \\
 & \ram & 0.1766 & 0.3079 & 0.1830 & 0.1092 \\
 & \hinge & 0.1876 & 0.3074 & 0.1961 & 0.1209 \\
 & \neuinfer & 0.1740 & 0.2808 & 0.1849 & 0.1153 \\
 & \gran & \underline{0.2901} & \underline{0.4020} & \underline{0.3019} & \underline{0.2310} \\
 & \ours & \color{black}{\textbf{0.3254}} & \color{black}{\textbf{0.5261}} & \color{black}{\textbf{0.3459}} & \color{black}{\textbf{0.2314}} \\
\midrule
\multirow{7}{*}{\hnfbs} & \nalp & 0.4578 & 0.6414 & 0.5160 & 0.3606 \\
 & \tnalp & 0.4357 & 0.6069 & 0.4865 & 0.3460 \\
 & \ram & 0.5788 & 0.6874 & 0.6085 & \color{black}{\textbf{0.5205}} \\
 & \hinge & 0.4809 & 0.6893 & 0.5412 & 0.3736 \\
 & \neuinfer & 0.3414 & 0.5334 & 0.3836 & 0.2466 \\
 & \gran & \color{black}{\textbf{0.5873}} & \underline{0.7223} & \underline{0.6286} & \color{black}{\underline{0.5149}} \\
 & \ours & \color{black}{\underline{0.5796}} & \color{black}{\textbf{0.7586}} & \color{black}{\textbf{0.6323}} & \color{black}{0.4857} \\
\bottomrule
\end{tabular}
\end{table}

\subsection{Link Prediction Results}
We show the performance of link prediction using standard metrics: Mean Reciprocal Rank (MRR), Hit@10, Hit@3, and Hit@1; higher values indicate better performance~\cite{survey}. Since \transea, \mtkgnn, \kbln, and \literale cannot consider qualifiers, we first consider the link prediction only on the primary triplets, i.e., we consider the case where the missing entity is only in a primary triplet. Table~\ref{tb:lp_tri} shows the link prediction results on the primary triplets, where the best results are boldfaced and the second-best results are underlined. We see that the transformer-based hyper-relational knowledge graph embedding methods, \stare, \hytrans, and \gran show better performance than the other baselines. More importantly, {\color{black} \ours significantly outperforms all the baseline methods in terms of all metrics on \hnwk and \hnyg. On \hnfbs, \ours outperforms all baselines in terms of \hten and \hthree, shows comparable performance to \ram in \mrr, and lags behind the best-performing baselines in \hone.} When predicting a missing entity in a primary triplet, the answer can be changed depending on its qualifiers. For example, Table~\ref{tb:lp_qual} shows examples of the predictions made by \ours on \hnwk where \ours successfully predicts the answers. 

We also consider the case where the missing entity can be placed either in a primary triplet or a qualifier. Table~\ref{tb:lp_all} shows the results in \hnwk and \hnfbs. In \hnyg, all qualifiers in the test set contain only numeric entities. Therefore, for \hnyg, the link prediction results on all entities of the hyper-relational facts are identical to those on the primary triplets. While \stare and \hytrans can handle the hyper-relational facts, their implementations are provided only for the prediction on the primary triplets but not for the qualifiers. Thus, we could not report the results of these methods. {\color{black}Table~\ref{tb:lp_all} shows that \ours achieves the best performance among all the methods in \hnwk in terms of all metrics, while \ours shows comparable performance to \gran in \hnfbs. The performance gap between \ours and the best baseline, \gran, is substantial in \hnwk. This is because \hnwk contains diverse numeric literals, and \gran does not effectively use that information, whereas \ours can appropriately interpret and utilize numeric literals.}

Recall that the link prediction task is to predict discrete entities but not numeric entities, according to our definition in Section~\ref{sec:task}. In Table~\ref{tb:lp_tri} and Table~\ref{tb:lp_all}, all methods solve the same problems where the missing entities are originally discrete entities. Even though we consider the numeric literals as discrete entities in \nalp, \tnalp, \ram, \hinge, \neuinfer, \stare, \hytrans, and \gran, we exclude the predictions on numeric literals when measuring the link prediction performance of these methods for a fair comparison. Instead, we use the predictions on numeric literals to measure the numeric value prediction performance of the aforementioned methods, which will be presented in Section~\ref{sec:nvp}.


\begin{table*}[t]
\small
\caption{Relation Prediction Results on the Primary Triplets (Tri) and All Relations of the Hyper-Relational Facts (All). Overall, \ours outperforms the baseline methods in relation prediction.}
\setlength{\tabcolsep}{0.84em}
\label{tb:rp}
\begin{tabular}{cc|cccc|cccc|cccc}
\Xhline{2\arrayrulewidth}
 & & \multicolumn{4}{c|}{\hnwk}  & \multicolumn{4}{c|}{\hnyg}  & \multicolumn{4}{c}{\hnfbs} \\
& & \mrr & \hten & \hthree & \hone & \mrr & \hten & \hthree & \hone & \mrr & \hten & \hthree & \hone \\
\Xhline{\arrayrulewidth}
 \multirow{6}{*}{Tri} & \nalp & 0.5682 & 0.6886 & 0.6014 & 0.5002 & 0.3419 & 0.5696 & 0.3926 & 0.2113 & 0.5539 & 0.6776 & 0.5889 & 0.4760 \\
&\tnalp & 0.5967 & 0.7542 & 0.6316 & 0.5164 & 0.4273 & 0.6840 & 0.4817 & 0.2957 & 0.5330 & 0.6873 & 0.5645 & 0.4528 \\
&\hinge & 0.8806 & 0.9700 & 0.9183 & 0.8295 & 0.8174 & \underline{0.9538} & 0.8786 & 0.7400 & 0.9684 & \underline{0.9964} & 0.9871 & 0.9483 \\
&\neuinfer & 0.7485 & 0.8929 & 0.7929 & 0.6734 & 0.6437 & 0.8965 & 0.7175 & 0.5162 & 0.7643 & 0.9427 & 0.8630 & 0.6523 \\
&\gran & \underline{0.9285} & \underline{0.9899} & \underline{0.9615} & \underline{0.8898} & \underline{0.8347} & 0.9383 & \underline{0.8858} & \underline{0.7697} & \textbf{0.9845} & 0.9947 & \underline{0.9934} & \textbf{0.9755} \\
&\ours & \color{black}{\textbf{0.9474}} & \color{black}{\textbf{0.9905}} & \color{black}{\textbf{0.9789}} & \color{black}{\textbf{0.9145}} & \textbf{0.8797} & \textbf{0.9851} & \color{black}{\textbf{0.9379}} & \textbf{0.8135} & \color{black}{\underline{0.9815}} & \color{black}{\textbf{0.9994}} & \color{black}{\textbf{0.9966}} & \color{black}{\underline{0.9669}} \\ 
\Xhline{\arrayrulewidth}
 \multirow{6}{*}{All} & \nalp & 0.7410 & 0.8245 & 0.7684 & 0.6927 & 0.4224 & 0.6222 & 0.4669 & 0.3078 & 0.7010 & 0.8306 & 0.7748 & 0.6084 \\
&\tnalp & 0.7625 & 0.8631 & 0.7912 & 0.7081 & 0.4946 & 0.7220 & 0.5441 & 0.3775 & 0.6545 & 0.8345 & 0.7471 & 0.5348 \\
&\hinge & 0.9278 & 0.9828 & 0.9527 & 0.8953 & 0.8397 & \underline{0.9594} & 0.8935 & 0.7718 & 0.9675 & \underline{0.9981} & 0.9924 & 0.9431 \\
&\neuinfer & 0.8320 & 0.9367 & 0.8800 & 0.7671 & 0.6803 & 0.9054 & 0.7479 & 0.5655 & 0.7980 & 0.9692 & 0.9098 & 0.6838 \\
&\gran & \underline{0.9599} & \underline{0.9941} & \underline{0.9784} & \underline{0.9382} & \underline{0.8548} & 0.9459 & \underline{0.8998} & \underline{0.7977} & \textbf{0.9918} & 0.9972 & \underline{0.9965} & \textbf{0.9870} \\
&\ours & \color{black}{\textbf{0.9706}} & \color{black}{\textbf{0.9947}} & \color{black}{\textbf{0.9881}} & \color{black}{\textbf{0.9522}} & \textbf{0.8944} & \textbf{0.9869} & \textbf{0.9455} & \textbf{0.8363} & \color{black}{\underline{0.9902}} & \color{black}{\textbf{0.9997}} & \color{black}{\textbf{0.9982}} & \color{black}{\underline{0.9825}} \\
\Xhline{2\arrayrulewidth}
\end{tabular}
\end{table*}

\subsection{Relation Prediction Results}
While all baseline methods provide the implementation of link prediction, relation prediction was only implemented in \nalp, \tnalp, \hinge, \neuinfer, and \gran. We compare the relation prediction performance of \ours with these methods. Table~\ref{tb:rp} shows the results of relation prediction on the primary triplets (Tri) and all relations of the hyper-relational facts (All). In \hnwk and \hnyg, we see that \ours achieves the highest \mrr, \hten, \hthree, and \hone. In \hnfbs, both \ours and \gran achieve over 99\% \hthree, indicating that these methods almost always correctly predict a missing relation within the top 3 predictions.



\begin{table}[t]
\small
\caption{Numeric Value Prediction Results on the Primary Triplets (Tri) and All Numeric Values in the Hyper-Relational Facts (All) in terms of RMSE ($\downarrow$).}
\setlength{\tabcolsep}{0.47em}
\label{tb:vp}
\begin{tabular}{ccccccc}
\toprule
 & \multicolumn{2}{c}{\hnwk} & \multicolumn{2}{c}{\hnyg} & \multicolumn{2}{c}{\hnfbs} \\
 & Tri & All & Tri & All & Tri & All \\
\midrule
\transea & 0.0772 & - & \underline{0.0778} & - & 0.1332 & - \\
\mtkgnn & 0.1390 & - & 0.1203 & - & 0.0908 & - \\
\kbln & 0.1550 & - & 0.1342 & - & 0.0890 & - \\
\literale & 0.2104 & - & 0.1783 & - & 0.0801 & - \\
\nalp & 0.2329 & 0.1681 & 0.1399 & 0.1375 & 0.1055 & 0.0894 \\
\tnalp & 0.2312 & 0.1673 & 0.1185 & 0.1176 & 0.0929 & 0.0923 \\
\ram & 0.1102 & \underline{0.0820} & 0.0969 & 0.1132 & 0.0706 & \underline{0.0627} \\
\hinge & 0.2435 & 0.1752 & 0.1141 & \underline{0.1123} & 0.1077 & 0.0939 \\
\neuinfer & 0.2425 & 0.1744 & 0.1176 & 0.1396 & 0.1093 & 0.1004 \\
\stare & 0.0832 & - & 0.0990 & - & 0.0670 & - \\
\hytrans & \underline{0.0761} & - & 0.0972 & - & \underline{0.0656} & - \\
\gran & 0.2293 & 0.1667 & 0.1180 & 0.1247 & 0.0835 & 0.0773 \\
\ours & \textbf{0.0548} & \textbf{0.0405} & \textbf{0.0706} & \textbf{0.0694} & \textbf{0.0532} & \textbf{0.0499} \\
\bottomrule
\end{tabular}
\end{table}

\begin{table}[htbp]
\small
\caption{Numeric Value Prediction Results per Attribute Type in \hnwk in terms of RMSE. \ours shows better performance than all other methods for each attribute type.}
\setlength{\tabcolsep}{0.68em}
\label{tb:vp_rel}
\begin{tabular}{ccccc}
\toprule
 & \multicolumn{3}{c}{Primary Triplet} & Qualifier \\
 & ranking & HDI & fertility rate & point in time \\
\midrule
Frequency & 11,951 & 708 & 353 & 18,613 \\
Minimum value & 1.0 & 0.190 & 0.827 & 1.0 \\
Maximum value & 246.0 & 0.957 & 7.742 & 2285.2 \\
Standard deviation & 49.2 & 0.166 & 1.594 & 35.8 \\
\midrule
\transea & 19.3 & 0.049 & \underline{0.279} & - \\
\mtkgnn & 40.4 & 0.079 & 0.569 & - \\
\kbln & 35.2 & 0.130 & 0.782 & - \\
\literale & 47.7 & 0.168 & 0.900 & - \\
\nalp & 68.2 & 0.074 & 0.760 & 94.4 \\
\tnalp & 67.1 & 0.065 & 1.149 & 81.8 \\
\ram & 22.7 & 0.066 & 1.472 & \underline{24.9} \\
\hinge & 70.5 & 0.077 & 1.363 & 91.8 \\
\neuinfer & 71.6 & 0.067 & 0.967 & 93.9 \\
\stare & 18.0 & 0.046 & 0.380 & - \\
\hytrans & \underline{16.0} & \underline{0.032} & 0.508 & - \\
\gran & 67.7 & 0.057 & 0.758 & 88.5 \\
\ours & \textbf{12.4} & \textbf{0.023} & \textbf{0.222} & \textbf{21.0} \\
\bottomrule
\end{tabular}
\end{table}

\begin{figure}
\centering
\begin{minipage}{\linewidth}
\centering
\subcaptionbox{Uruguay}
	{\includegraphics[width=0.48\linewidth]{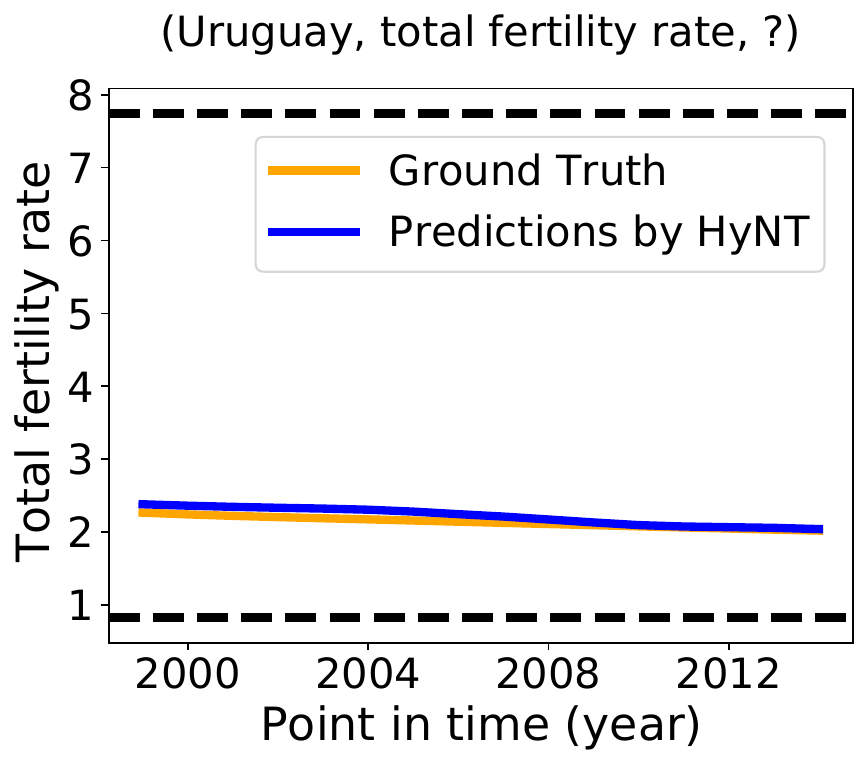}}
\subcaptionbox{Ghana}
	{\includegraphics[width=0.48\linewidth]{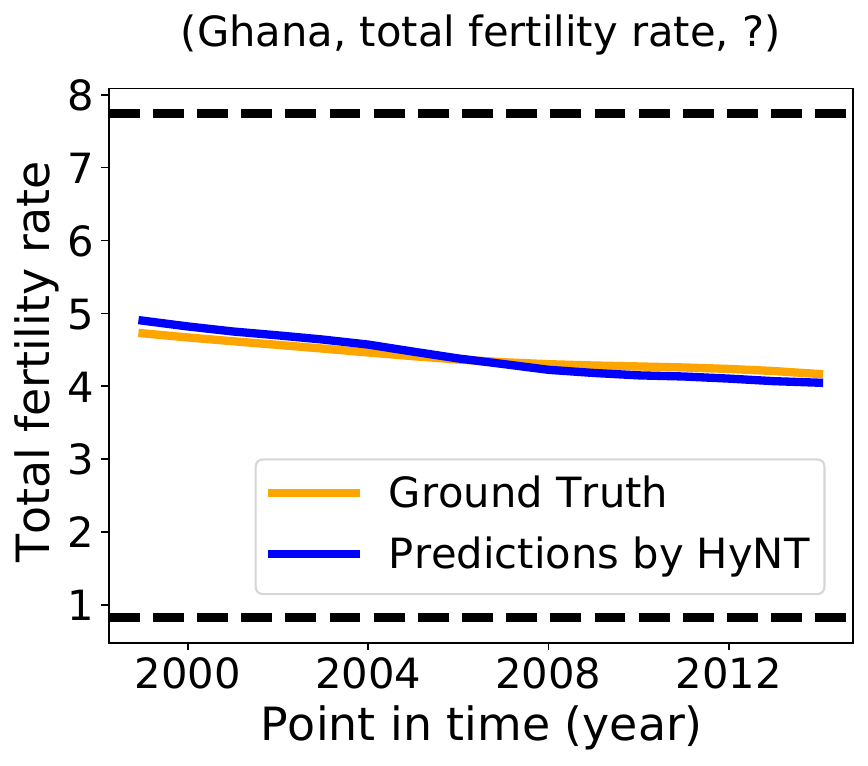}}
\caption{Comparison between \ours's predictions and the ground-truth values of the total fertility rates of Uruguay and Ghana over time.}
\label{fig:plot1}
\end{minipage}

\begin{minipage}{\linewidth}
\centering
\subcaptionbox{Sweden National Team}
	{\includegraphics[width=0.48\linewidth]{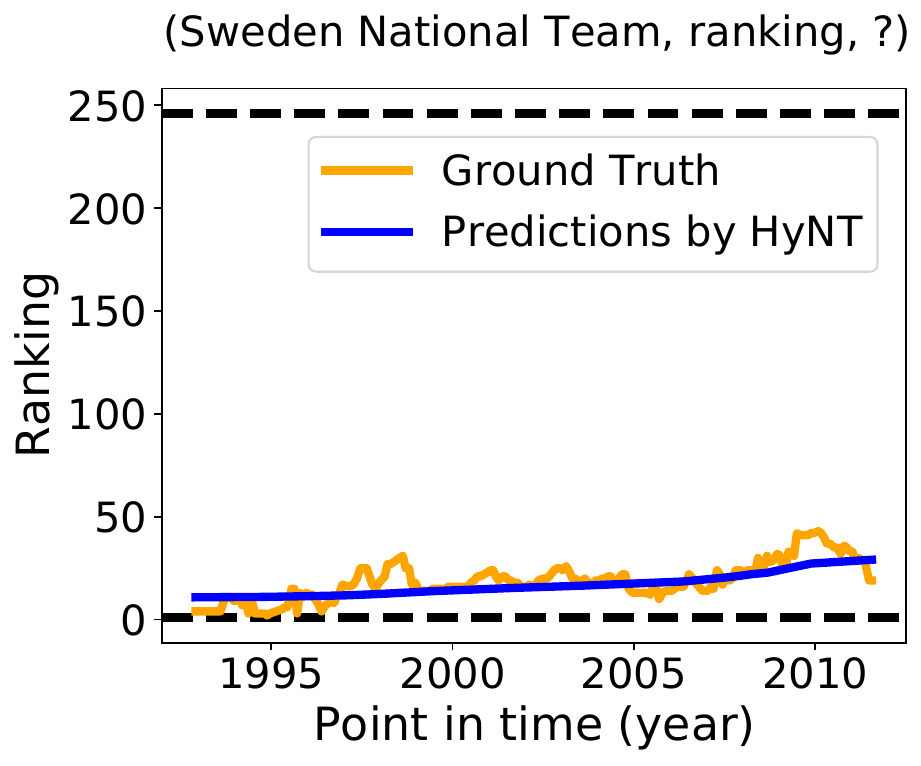}}
\subcaptionbox{India National Team}
	{\includegraphics[width=0.48\linewidth]{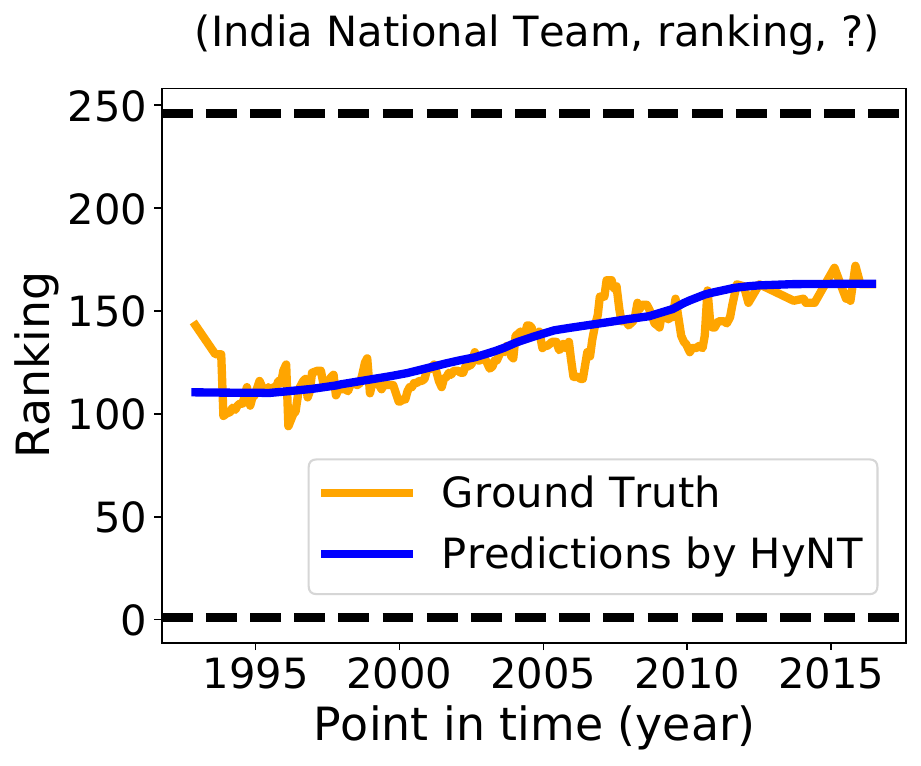}}
\caption{Comparison between \ours's predictions and the ground-truth values of rankings of Sweden National Team and India National Team over time.}
\label{fig:plot2}
\end{minipage}
\end{figure}


\subsection{Numeric Value Prediction Results}
\label{sec:nvp}
Table~\ref{tb:vp} shows the numeric value prediction results in terms of the Root-Mean-Square Error (RMSE). The lower RMSE indicates better performance. Since the scales of the numeric values vary depending on the types of attributes, e.g., the running time of a TV episode varies from 10 minutes to 360 minutes, whereas the number of days in a month varies from 28 days to 31 days; we apply the min-max normalization for each attribute type to rescale all numeric values from 0 to 1. Since \transea, \mtkgnn, \kbln, and \literale cannot handle hyper-relational facts, we first measure the performance on the primary triplets where the missing values are placed only in the primary triplets (denoted by Tri). We also consider the case where the missing numeric values can be either on a primary triplet or a qualifier (denoted by All). Note that \stare and \hytrans do not provide the implementations of the prediction on the qualifiers. In Table~\ref{tb:vp}, we see that \ours achieves the least RMSE on all datasets in all settings. We further analyze the results by selecting a few attribute types among the diverse numeric literals. Table~\ref{tb:vp_rel} shows the numeric value prediction performance on `ranking', `Human Development Index (HDI)', `fertility rate', and `point in time' where we present the raw data without rescaling. For `point in time', we convert the date into a real number, e.g., converting January 28th, 1922, into 1922 + 28/365. In Table~\ref{tb:vp_rel}, we also present the frequency, the minimum, the maximum, and the standard deviation of each attribute type. We see that \ours shows the best performance among all the methods. We also consider a numeric value prediction problem in the form of $((h, r, ?), \{(\mathrm{point\ in\ time}, v_1)\})$, $((h, r, ?), \{(\mathrm{point\ in\ time}, v_2)\})$, $\cdots$ in \hnwk where $v_1, v_2, \cdots$ correspond to different points in time. Let us focus on the missing numeric values in the primary triplet and call them target values. By visualizing the target values associated with different qualifiers indicating different points in time, we trace how the target values change over time. In Figure~\ref{fig:plot1}, we show \ours's predictions and the ground-truth values of the total fertility rate of Uruguay and Ghana over time. In Figure~\ref{fig:plot2}, we show the rankings of Sweden National Team and India National Team over time. The dotted lines indicate the minimum and the maximum values. We see that the predictions made by \ours are very close to the ground-truth numeric values.

\subsection{Ablation Studies of \ours}
Table~\ref{tb:ab_num} shows the link prediction and numeric value prediction results of \ours for all entities and numeric values of the hyper-relational facts in \hnwk where we treat the numeric values as discrete entities (denoted by Discrete), or we treat them as numeric entities (denoted by Numeric). As described in Section~\ref{sec:base}, a straightforward way to handle numeric entities in a hyper-relational knowledge graph is to treat them as discrete entities like how we feed the numeric entities into the baseline methods that are not designed for numeric literals. In Table~\ref{tb:ab_num}, we see that the way we handle numeric values in \ours is much more effective, leading to better performance on both link prediction and numeric value prediction.

Table~\ref{tb:ab} shows the ablation studies of \ours on \hnwk. We analyze MRR scores of link and relation predictions and RMSE of numeric value prediction. We consider the case where the predictions are made only on the primary triplets (Tri) and the case where the predictions are made on all components of the hyper-relational facts (All). In Section~\ref{sec:loss}, we introduce the masking strategy where we mask a component in a hyper-relational fact. The first five rows in Table~\ref{tb:ab} indicate the cases where we do not mask some types of components. We see that if we do not mask relations ($\sR$), the relation prediction performance significantly degrades. Similarly, if we do not mask numeric entities ($\sV_\text{N}$), the numeric value prediction performance degrades. Also, if we do not mask qualifiers ($\sE_\text{qual}$), `All' performances degrade because the model is not trained to perform predictions on the qualifiers. On the other hand, we also replace a prediction transformer with a simple linear projection layer (denoted by w/o prediction transformer). Lastly, we change the triplet encoding and the qualifier encoding in Section~\ref{sec:aggr} by replacing them with the Hadamard product, i.e., we compute $\vx_\mathrm{tri} = \vh \circ \vr \circ \vt$ and $\vx_{\mathrm{qual}_i} = \vq_i \circ \vecv_i$ (denoted by w/ Hadamard encoding). While the last two variations are not as critical as removing maskings, \ours achieves the best performance when all pieces are together.

\begin{table}[t]
\small
\caption{Link Prediction and Numeric Value Prediction Using Different Handling of Numeric Entities in \ours on \hnwk.}
\setlength{\tabcolsep}{0.55em}
\label{tb:ab_num}
\begin{tabular}{ccccccc}
\toprule
 & \mrr($\uparrow$) & \hten($\uparrow$) & \hthree($\uparrow$) & \hone($\uparrow$) & \rmse($\downarrow$) \\
\midrule
Discrete & \color{black}{0.2478} & \color{black}{0.3713} & \color{black}{0.2621} & \color{black}{0.1827} & 0.2267 \\
Numeric & \color{black}{\textbf{0.3254}} & \color{black}{\textbf{0.5259}} & \color{black}{\textbf{0.3459}} & \color{black}{\textbf{0.2314}} & \textbf{0.0405} \\
\bottomrule
\end{tabular}
\end{table}

\begin{table}[t]
\footnotesize
\caption{Link/Relation/Numeric Value Predictions on the Primary Triplets (Tri) and All Components in Hyper-relational facts (All) for Ablation Studies of \ours on \hnwk.}
\setlength{\tabcolsep}{4.4pt}
\label{tb:ab}
\begin{tabular}{l*{6}{p{6mm}}}
\toprule
 & \multicolumn{2}{@{\hspace{0.3\tabcolsep}}c@{\hspace{0.3\tabcolsep}}}{Link(\mrr\hspace{-1pt}$\uparrow$)} & \multicolumn{2}{@{\hspace{0.3\tabcolsep}}c@{\hspace{0.3\tabcolsep}}}{Relation(\mrr\hspace{-1pt}$\uparrow$)} & \multicolumn{2}{@{\hspace{0.3\tabcolsep}}c@{\hspace{0.3\tabcolsep}}}{Value(\rmse\hspace{-1pt}$\downarrow$)}\\
 & \hfil Tri & \hfil All & \hfil Tri & \hfil All & \hfil Tri & \hfil All \\
\midrule
w/o masking $\sV_\text{N}, \sR, \sE_\text{qual}$ & \hfil \color{black}{0.277} & \hfil \color{black}{0.264} & \hfil \color{black}{0.014} & \hfil \color{black}{0.018} & \hfil 0.507 & \hfil 0.733 \\
w/o masking $\sR, \sE_\text{qual}$ & \hfil \color{black}{0.291} & \hfil \color{black}{0.278} & \hfil \color{black}{0.074} & \hfil \color{black}{0.043} & \hfil 0.064 & \hfil 0.112 \\
w/o masking $\sV_\text{N}$ & \hfil \color{black}{0.291} & \hfil \color{black}{0.312} & \hfil \color{black}{0.939} & \hfil \color{black}{0.966} & \hfil 0.429 &\hfil 0.786 \\
w/o masking $\sR$ & \hfil \color{black}{0.259} & \hfil \color{black}{0.272} & \hfil \color{black}{0.015} & \hfil \color{black}{0.010} & \hfil 0.059 & \hfil 0.042 \\
w/o masking $\sE_\text{qual}$ & \hfil \color{black}{0.256} & \hfil \color{black}{0.244} & \hfil \color{black}{0.936} & \hfil \color{black}{0.514} & \hfil 0.064 & \hfil 0.329 \\
w/o prediction transformer & \hfil \color{black}{0.275} & \hfil \color{black}{0.296} & \hfil \color{black}{0.930} & \hfil \color{black}{0.961} & \hfil 0.060 & \hfil 0.043 \\
w/ Hadamard encoding & \hfil \color{black}{0.258} & \hfil \color{black}{0.255} & \hfil \color{black}{0.939} & \hfil \color{black}{0.964} & \hfil 0.064 &\hfil 0.048 \\
$\ours$ & \hfil \color{black}{0.304} & \hfil \color{black}{0.325} & \hfil \color{black}{0.947} & \hfil \color{black}{0.971 }& \hfil 0.055 & \hfil 0.040 \\
\bottomrule
\end{tabular}
\end{table}

\section{Conclusion \& Future Work}
We propose \ours which is a transformer-based representation learning method for hyper-relational knowledge graphs having various numeric literals. One of the key ideas is to learn representations of a primary triplet and each qualifier by allowing them to exchange information with each other, where the relative importance is also learned using an attention mechanism. \ours considers link prediction, numeric value prediction, and relation prediction tasks to compute embeddings of entities and relations, where entities can be either discrete or numeric. Experiments show that \ours significantly outperforms 12 different state-of-the-art methods.

We will extend \ours to an inductive learning setting which assumes that some entities and relations can only appear at test time~\cite{qblp,ingram}. Also, we plan to consider incorporating text descriptions or images into \ours. Additionally, we will explore how \ours can be utilized in various domains, such as conditional link prediction or triplet prediction~\cite{bive} and applications requiring advanced knowledge representation~\cite{robokg}.


\begin{acks}
This research was partly supported by NRF grants funded by MSIT (2022R1A2C4001594 and 2018R1A5A1059921). This work was also supported by IITP grants funded by MSIT 2022-0-00369 (Development of AI Technology to support Expert Decision-making that can Explain the Reasons/Grounds for Judgment Results based on Expert Knowledge) and 2020-0-00153 (Penetration Security Testing of ML Model Vulnerabilities and Defense). 
\end{acks}

\bibliographystyle{ACM-Reference-Format}
\balance
\bibliography{hynt_ref_camera}
\clearpage
\appendix
\section*{Appendix}
\section{Details about Datasets}
\label{app:data}
We create three real-world {\hnkg}s, named \hnwk, \hnyg, and \hnfb. Similar to~\cite{mmkg,stare}, we construct the datasets consisting of the entities in~\fbk~\cite{fb237} and their related numeric values. For~\hnwk, we first extract all hyper-relational facts about the entities in \wiki~\cite{wk} that are mapped from~\fbk. Then, we filter the triplets and qualifiers so that the remaining entities are either from~\fbk or numeric values. For~\hnyg, since there is no direct mapping from~\fb~\cite{fb} to~\yago~\cite{yago}, we first map~\fbk entities to~\dbp~\cite{db}, and map the converted entities to~\yagot~\cite{yagot}. We extract the triplets whose head and tail entities match~\fbk entities and also extract the numeric literals of the extracted entities. Then, the qualifiers of the extracted triplets are added if the entities of the qualifiers match the \fbk entities or the entities in the qualifiers are numeric literals. For \hnfb, we first transform Compound Value Types (CVTs) in \fb into hyper-relational facts by following \cite{fbtw}. Similar to~\hnwk, we filter the triplets and qualifiers so that the remaining entities are either from~\fbk or numeric literals. When we create our datasets, we remove inverse and near-duplicate relations to prevent data leakage problems by following~\cite{fb237}.

We have inspected the minimum and the maximum values of numeric entities per relation and found that the raw data include some incorrect information. We have filtered out those incorrect triplets or qualifiers. For example, the triplet \textsf{(Glee, number of seasons, 91)} was removed since Glee has a total of 6 seasons. In the process of removing the incorrect triplets, 0.04\% (\hnwk), 0.52\% (\hnyg), and 0.02\% (\hnfb) of triplets were removed.


\section{Details about Baseline Methods}
\label{app:base}
For the hyperparameters of the baseline methods, we follow the notation from the original papers. For all models, we set the embedding dimension to 256 unless otherwise stated. We conduct our experiments using GeForce RTX 2080 Ti, GeForce RTX 3090, or RTX A6000 by considering the dependencies of the implementations of each method. We run \nalp, \tnalp, \hinge, \neuinfer, \stare, \hytrans, and \gran using GeForce RTX 2080 Ti, and run \ram using GeForce RTX 3090 and RTX A6000, and run \transea, \mtkgnn, \kbln, \literale, and \ours using GeForce RTX 2080 Ti, GeForce RTX 3090, and RTX A6000.

\paragraph*{\bf \transea~\cite{transea}}
We re-implemented \transea using PyTorch since the original code has inconsistencies with the description in the paper: in the original code, the bias term is not computed per attribute. We perform a grid search on $\lambda=\{0.1, 0.5, 1.0\}$, $\gamma=\{2.5, 5.0, 7.5, 10.0\}$, and $\alpha=\{0.25, 0.5, 0.75\}$. We perform validation every 50 epochs up to 500 epochs and select the best hyperparameters using the validation set.

\paragraph*{\bf \mtkgnn~\cite{mtkgnn}}
Since the official implementation of \mtkgnn is not provided, we use the implementation of \mtkgnn provided in \cite{literale}. We modified the implementation so that the regression loss does not take the entities without numeric literals into account, as stated in the original paper. We tune the model with the learning rate $=\{0.0001, 0.001\}$, $d=\{0.1, 0.2\}$, and the label smoothing ratio $=\{0.0, 0.1\}$ for all datasets. We train the model for 100 epochs where we conduct validation every epoch and select the best epoch.

\paragraph*{\bf \kbln~\cite{kblrn}}
In the original \kblrn~\cite{kblrn}, the module for relational features requires extra logical rules. Thus, by following~\cite{literale}, we use \kbln provided in~\cite{literale} instead of \kblrn for fair comparison. We try the learning rate $=\{0.0001, 0.001, 0.01\}$, the dropout rate $=\{0.1, 0.2\}$ and the label smoothing ratio $=\{0.0, 0.1\}$ for all datasets. We first train the model for 100 epochs on a link prediction task, where we perform validation every epoch and select the best epoch. Then, we additionally train a linear regression layer for a numeric value prediction task by following \cite{mtkgnn}. For the training of the linear regression layer, we try the learning rate $=\{0.0005, 0.01\}$ and the number of epochs $=\{100, 500\}$ for all datasets.

\paragraph*{\bf \literale~\cite{literale}}
Similar to \kbln, we first train \literale for 100 epochs on a link prediction task and then additionally train a linear regression layer for a numeric value prediction. We tune \literale with the learning rate $=\{0.001, 0.01, 0.1\}$, the embedding dropout rate $=\{0.1, 0.2\}$, the feature map dropout rate $=\{0.1, 0.2\}$, and the projection layer dropout rate $=\{0.2, 0.3\}$. For the training of the linear regression layer, we try the learning rate $=\{0.0005, 0.01\}$ and the number of epochs $=\{100, 500\}$ for all datasets.

\paragraph*{\bf \nalp~\cite{nalp}}
Since the official implementation of \nalp saves all permutations of qualifiers in the pre-processing step, it is not scalable enough to process our datasets; thus, we re-implemented \nalp. This also applies to \tnalp, \hinge, and \neuinfer since they share the same implementation for the pre-processing step. On all datasets, we try $\lambda=\{ 0.00001, 0.00005, 0.0001 \}$, $n_f=\{ 100, 200 \}$, and $n_\mathrm{gFCN}=\{ 800, 1000, 1200 \}$. We perform validation every 500 epochs up to 5,000 epochs and select the best hyperparameters.

\paragraph*{\bf \tnalp~\cite{tnalp}}
On all datasets, we try $\lambda=\{ 0.00001, 0.00005, 0.0001 \}$, $n_f=\{ 100, 200 \}$, $n_\mathrm{gFCN}=\{ 800, 1000, 1200 \}$, and $n_\mathrm{tFCN}=\{ 100, 200 \}$. We perform validation every 500 epochs up to 5,000 epochs.

\paragraph*{\bf \ram~\cite{ram}}
Since the total embedding dimension of entities in \ram is $d \times m$, we use $d = 128$ and $m = 2$ for \hnyg and \hnfbs. However, due to GPU memory constraints, we use $d = 32$ and $m = 2$ for \hnwk, \wikipm, and \wdk. We tune \ram with the learning rate $=\{0.001, 0.002, 0.003, 0.005\}$, the decay rate $=\{0.99, 0.995\}$, and the dropout rate $=\{0.0, 0.2, 0.4\}$ for all datasets. We run \ram with a maximum of 200 epochs, with an early stopping strategy adopted. For \hnyg and \wikipm, we perform validation every epoch with the validation patience of 10, while for \hnwk, \wdk, and \hnfbs, we perform validation every 5 epochs with the same validation patience.

\paragraph*{\bf \hinge~\cite{hinge}}
We try $\lambda=\{ 0.00001, 0.00005, 0.0001 \}$ and $n_f=\{ 100, 200, 400 \}$ for all datasets. We perform validation every 100 epochs up to 1,000 epochs. 

\paragraph*{\bf \neuinfer~\cite{neuinfer}}
We try $\lambda=\{ 0.00001, 0.00005, 0.0001 \}$, $d=\{ 800, 1000, 1200 \}$, $n_1=\{ 1, 2 \}$, $n_2=\{ 1, 2 \}$, and $w=\{ 0.1, 0.3 \}$ for all datasets. We perform validation every 500 epochs up to 5,000.

\paragraph*{\bf \stare~\cite{stare}}
We tune \stare with the number of \stare layers $=\{1, 2\}$, the number of transformer layers $=\{1, 2\}$, the \stare dropout rate $=\{0.2, 0.3\}$, and the transformer dropout rate $=\{0.1, 0.2\}$ for all datasets. We run \stare for 400 epochs and perform validation every 5 epochs.

\paragraph*{\bf \hytrans~\cite{hytrans}}
We tune \hytrans with the number of transformer layers $=\{1, 2\}$, the qualifier perturb probability $=\{0.0, 0.5, 1.0\}$, the transformer dropout rate $=\{0.1, 0.2\}$, and the entity embedding matrix dropout rate $=\{0.2, 0.3\}$ for all datasets. We run \hytrans for 400 epochs and perform validation every 10 epochs.

\paragraph*{\bf \gran~\cite{gran}}
We use $b = 256$ for \hnwk due to GPU memory constraints. We try $\rho=\{ 0.1, 0.2, 0.3 \}$, $\epsilon^{(e)}=\{ 0.2, 0.4, 0.6, 0.8 \}$, and $\epsilon^{(r)}=\{ 0.0, 0.2, 0.4 \}$ for all datasets. We perform validation every 20 epochs up to 200 epochs.

\section{Experimental Results on \hnfb}
\label{app:fb}
As explained in Section~\ref{sec:base}, among 12 baseline methods, all methods handling hyper-relational facts (i.e., \nalp, \tnalp, \ram, \hinge, \neuinfer, \stare, \hytrans, and \gran) fail to process \hnfb since they are not scalable enough to process this scale of data. The remaining four baseline methods (i.e., \transea, \mtkgnn, \kbln, and \literale) are the methods dealing with numeric literals but not hyper-relational facts. Note that \ours is the only method that can handle both numeric literals and hyper-relational facts and can process \hnfb.

Table~\ref{tb:fb_all} shows the link prediction and numeric value prediction results of \ours on \hnfb. We report the prediction performance on the primary triplets (Tri) and all entities of the hyper-relational facts (All). The best results are boldfaced and the second-best results are underlined. We see that \ours significantly outperforms all the baseline methods in terms of all metrics. Also, Table~\ref{tb:fb_all_rp} shows the results of relation prediction on the primary triplets (Tri) and all relations of the hyper-relational facts (All) on \hnfb. Since \transea, \mtkgnn, \kbln, and \literale do not provide implementations for relation prediction, we could not report the relation prediction results of these methods. Even though we cannot compare \ours with other methods on \hnfb, we see that \ours shows around 97\% \hone on \hnfb, indicating that \ours correctly predicts missing relations with a 97\% chance.

\section{Experimental Results on \wikipm and \wdk}
\label{app:bench}
In Section~\ref{sec:related}, we discuss the well-known benchmark datasets for hyper-relational knowledge graphs, \jf~\cite{mtransh,jf}, \wdk~\cite{stare}, \wikipm~\cite{hinge}, and \wikip~\cite{nalp}. Among these, the \jf dataset has been considered problematic because it includes many redundant entities that cause data leakage problems. In Table~\ref{tb:lp_bmk}, we show link prediction performance on \wikipm and \wdk which include hyper-relational facts containing only discrete entities but not numeric entities. While \stare, \hytrans, and \gran retrained their methods using both training and validation sets in their original papers, we trained all methods only using the training set to be consistent with our other experiments. In Table~\ref{tb:lp_wd}, we show link prediction performance on \wdk after retraining using both training and validation sets. In both tables, we see that \ours shows better or comparable performance to the best baseline methods depending on the metric and the dataset. We provide the implementations, hyperparameters, and checkpoints of \ours at \url{https://github.com/bdi-lab/HyNT}.

\begin{table}[b]
\small
\caption{Link Prediction and Numeric Value Prediction Results on \hnfb. Our model, \ours, significantly outperforms all baseline methods in terms of all metrics.}
\setlength{\tabcolsep}{0.3em}
\label{tb:fb_all}
\begin{tabular}{ccccccc}
\toprule
& & \mrr($\uparrow$) & \hten($\uparrow$) & \hthree($\uparrow$) & \hone($\uparrow$) & \rmse($\downarrow$) \\
\midrule
\multirow{5}{*}{Tri} & \transea & 0.2327 & \underline{0.4793} & \underline{0.2894} & 0.1084 & \underline{0.0637} \\
& \mtkgnn & 0.2393 & 0.4270 & 0.2578 & 0.1540 & 0.2072 \\
& \kbln & 0.2343 & 0.4204 & 0.2523 & 0.1494 & 0.0735 \\
& \literale & \underline{0.2602} & 0.4559 & 0.2872 & \underline{0.1695} & 0.0678 \\
& \ours & \color{black}{\textbf{0.4544}} & \color{black}{\textbf{0.6571}} & \color{black}{\textbf{0.5036}} & \color{black}{\textbf{0.3520}} & \textbf{0.0517} \\
\midrule
All & \ours & \color{black}{\textbf{0.5022}} & \color{black}{\textbf{0.7207}} & \color{black}{\textbf{0.5561}} & \color{black}{\textbf{0.3939}} & \textbf{0.0558} \\
\bottomrule
\end{tabular}
\end{table}

\begin{table}[b]
\caption{Relation Prediction Results of \ours on \hnfb.}
\label{tb:fb_all_rp}
\begin{tabular}{ccccc}
\toprule
& \mrr & \hten & \hthree & \hone \\
\midrule
Tri & 0.9809 & 0.9990 & 0.9958 & 0.9662 \\
\midrule
All & 0.9860 & 0.9987 & 0.9954 & 0.9766 \\
\bottomrule
\end{tabular}
\end{table}

\begin{table}[b]
\footnotesize
\caption{Link Prediction on \wikipm and \wdk obtained by training the models only using the training set.}
\setlength{\tabcolsep}{0.55em}
\label{tb:lp_bmk}
\begin{tabular}{cccccccc}
\toprule
 & & \multicolumn{3}{c}{\wikipm} & \multicolumn{3}{c}{\wdk} \\
 & & \mrr & \hten & \hone & \mrr & \hten & \hone \\
\midrule
\multirow{9}{*}{Tri} & \nalp & 0.3563 & 0.4994 & 0.2713 & 0.2303 & 0.3471 & 0.1697 \\
& \tnalp & 0.3577 & 0.4857 & 0.2876 & 0.2205 & 0.3308 & 0.1630 \\
& \ram & 0.4593 & 0.5837 & \underline{0.3843} & 0.2756 & 0.3988 & 0.2104 \\
& \hinge & 0.3929 & 0.5467 & 0.3092 & 0.2636 & 0.4099 & 0.1870 \\
& \neuinfer & 0.3568 & 0.5326 & 0.2469 & 0.2202 & 0.3466 & 0.1543 \\
& \stare & 0.4579 & \textbf{0.6108} & 0.3640 & 0.3087 & 0.4515 & 0.2340 \\
& \hytrans & 0.4597 & 0.5942 & 0.3818 & 0.3041 & 0.4427 & 0.2309 \\
& \gran & \underline{0.4616} & \underline{0.6097} & 0.3664 & \underline{0.3299} & \underline{0.4722} & \underline{0.2551} \\
& \ours & \textbf{0.4820} & 0.6020 & \textbf{0.4153} & \textbf{0.3332} & \textbf{0.4743} & \textbf{0.2594} \\
\midrule
\multirow{7}{*}{All} & \nalp & 0.3601 & 0.5034 & 0.2751 & 0.2508 & 0.3747 & 0.1866 \\
& \tnalp & 0.3607 & 0.4896 & 0.2900 & 0.2425 & 0.3597 & 0.1815 \\
& \ram & 0.4606 & 0.5850 & \underline{0.3856} & 0.2955 & 0.4156 & 0.2316 \\
& \hinge & 0.3947 & 0.5488 & 0.3108 & 0.2771 & 0.4244 & 0.1996 \\
& \neuinfer & 0.3569 & 0.5324 & 0.2475 & 0.2254 & 0.3549 & 0.1581 \\
& \gran & \underline{0.4654} & \textbf{0.6133} & 0.3706 & \textbf{0.3605} & \textbf{0.5013} & \underline{0.2863} \\
& \ours & \textbf{0.4811} & \underline{0.6025} & \textbf{0.4139} & \underline{0.3599} & \underline{0.4999} & \textbf{0.2867} \\
\bottomrule
\end{tabular}
\end{table}

\begin{table}[b]
\caption{ Link Prediction on \wdk obtained by training the models using both training and validation sets. Baseline results are either from their original papers or from \stare~\cite{stare}.}
\label{tb:lp_wd}
\begin{tabular}{ccccc}
\toprule
 & & \multicolumn{3}{c}{\wdk} \\
 & & \mrr & \hten & \hone \\
\midrule
\multirow{5}{*}{Tri} & \nalp & 0.1770 & 0.2640 & 0.1310 \\
& \hinge & 0.2430 & 0.3770 & 0.1760 \\
& \stare & 0.3490 & 0.4960 & 0.2710 \\
& \hytrans & \underline{0.3560} & \underline{0.4980} & \underline{0.2810} \\
& \ours & \textbf{0.3568} & \textbf{0.5010} & \textbf{0.2811} \\
\midrule
\multirow{1}{*}{All} & \ours & \textbf{0.3834} & \textbf{0.5267} & \textbf{0.3081} \\
\bottomrule
\end{tabular}
\end{table}

\end{document}